%% file: main.tex
\documentclass[10pt,twocolumn,letterpaper]{article}

\usepackage[pagenumbers]{cvpr} 

\input{preamble}
\definecolor{cvprblue}{rgb}{0.21,0.49,0.74}
\usepackage[pagebackref,breaklinks,colorlinks,allcolors=cvprblue]{hyperref}
\usepackage{xspace}
\usepackage{threeparttable}
\usepackage{multirow}
\usepackage{caption}
\usepackage{subcaption}
\usepackage{wrapfig}
\usepackage{booktabs}
\usepackage{pifont}
\usepackage{makecell}
\usepackage{enumitem}
\usepackage{marvosym}
\usepackage{adjustbox}
\usepackage{nicematrix}

\usepackage{tcolorbox}
\newcommand{\name}{CodeDance\xspace}

\newcommand{\rname}{BAT\xspace}

\usepackage{mwe}

\usepackage{tcolorbox}
\usepackage{xcolor,colortbl}
\colorlet{lightpink}{pink!35}
\colorlet{lightcyan}{cyan!20}
\colorlet{lightgray}{gray!15}
\definecolor{darkgray}{rgb}{0.9, 0.9, 0.9}
\definecolor{lightgreen}{rgb}{0.886, 0.941, 0.851}
\definecolor{Rubinered}{rgb}{0.8, 0.0, 0.2}

\renewcommand{\cite}{\citep}
\definecolor{mincolor}{rgb}{0,0.6,0}
\definecolor{maxcolor}{rgb}{0.8,0,0}
\usepackage[dvipsnames]{xcolor}

\usepackage{lipsum}
\usepackage{multirow}
\usepackage{xcolor,colortbl}
\usepackage{graphicx}
\usepackage{wrapfig}
\usepackage{caption}
\usepackage{svg}
\definecolor{lightgreen}{rgb}{0.886, 0.941, 0.851}

\usepackage{enumitem}
\usepackage{xspace}

\def\paperID{15290} 
\def\confName{CVPR}
\def\confYear{2026}

\title{
CodeDance: A Dynamic Tool-integrated MLLM for \\ Executable Visual Reasoning  
}
\author{Qi Song\textsuperscript{1,4}{\thanks{Equal contribution. ${}^\dagger$Corresponding authors.
}
} \quad
Honglin Li\textsuperscript{2,4}${}^*$ \quad
Yingchen Yu\textsuperscript{3} \quad
Haoyi Zhou\textsuperscript{1} \quad
Lin Yang\textsuperscript{2} \quad \\
Song Bai\textsuperscript{3} \quad
Qi She\textsuperscript{4} \quad
Zilong Huang\textsuperscript{3}${}^\dagger$ \quad
Yunqing Zhao\textsuperscript{3}${}^\dagger$ \quad \\
\textsuperscript{1}Beihang University \quad
\textsuperscript{2}Westlake University \\
\textsuperscript{3}ByteDance Singapore \quad
\textsuperscript{4}ByteDance China\\
{\tt\small zilong.huang2020@gmail.com}
\quad
{\tt\small yunqing.z.0817@gmail.com}
\\
\href{https://codedance-vl.github.io/}{\textcolor{RubineRed} {CodeDance-VL.github.io}}
\vspace{-1mm}
}

\begin{document}
\maketitle
\input{sec/0_abstract}    
\input{sec/1_intro}

\input{sec/2_RelatedWorks}
\input{sec/3_method}

\input{sec/4_exps}
\input{camera_ready/sec/5_discussion}

\section*{Acknowledgements}
This work was supported by the grants from the Natural Science Foundation of China (62202029), and Young Elite Scientists Sponsorship Program by CAST (No. 2023QNRC001).

\input{camera_ready/sec/X_suppl}

\clearpage
\clearpage
{
    \small
    \bibliographystyle{ieeenat_fullname}
    \bibliography{main}
}

\end{document}

%% file: sec/0_abstract.tex
\begin{abstract}
Recent releases such as o3 highlight human-like “thinking with images” reasoning that combines tool use with stepwise verification, yet most open-source approaches still rely on text-only chains, rigid visual schemas, or single-step pipelines, limiting flexibility, interpretability, and transferability on complex tasks.
We introduce \textbf{\name}, which explores executable code as a general solver for visual reasoning. Unlike fixed-schema calls (e.g., only predicting bounding-box coordinates), \name defines, composes, and executes code to orchestrate multiple tools, compute intermediate results, and render visual artifacts (e.g., boxes, lines, plots) that support transparent, self-checkable reasoning.
To guide this process, we introduce a reward for balanced and adaptive tool calling, which balances exploration with efficiency and mitigates tool overuse.
Interestingly, beyond the expected capabilities taught by atomic supervision, we empirically observe novel emergent behaviors during RL training: \name demonstrates novel tool invocations, unseen compositions, and cross-task transfer. 
These behaviors arise without task-specific fine-tuning, suggesting a general and scalable mechanism for executable visual reasoning.
Extensive experiments across reasoning benchmarks (e.g., visual search, math, chart QA) show that \name not only consistently outperforms schema-driven and text-only baselines, but also surpasses closed models such as GPT-4o and larger open-source models.
\end{abstract}

%% file: sec/1_intro.tex
\begin{figure*}[t] 
\newcommand{\colwid}{0.32\linewidth} 
    \centering 
    \includegraphics[width=0.92\linewidth]{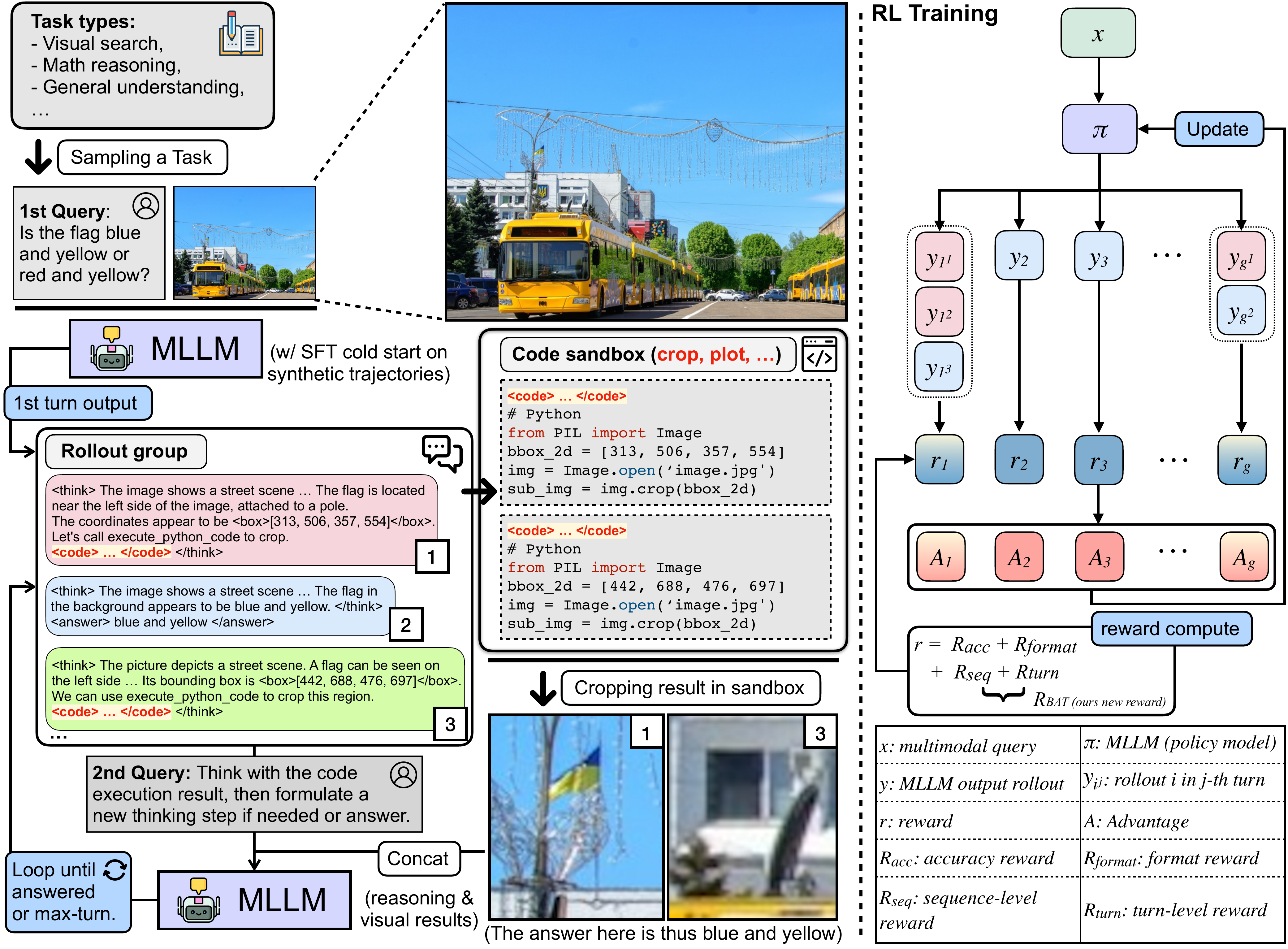}
    \vspace{-2mm}
    \caption{\textbf{Overview of our framework} that enables executable visual reasoning and invokes tool integration adaptively.
    }
    \label{fig:method} 
\end{figure*}

\section{Introduction}
\label{sec:introduction}

Multimodal Large Language Models (MLLMs) have made rapid progress, showing strong capabilities in both visual perception and reasoning. By leveraging the language-centric chain-of-thought (CoT) mechanism~\cite{brown2020language,wei2022chain}, models can decompose complex problems into intermediate steps, thereby improving performance on challenging tasks. However, when extended to visual tasks, traditional CoT relies on static context, preventing models from dynamically interacting with visual inputs or incorporating new observations during intermediate reasoning~\cite{zou2024look,chung2025don}, creating an information bottleneck that hinders multi-round focusing and validation. To address this, the o3 system~\cite{openai2025o3} integrates the ability to actively seek new information through multiple tool invocations, supporting iterative reasoning over visual inputs and demonstrating strong perception and analysis.


\begin{table*}[t]
\caption{Experiment results on benchmarks including Counting, Visual Search, and General datasets. $^\dagger$ denotes reproduced results.}
\label{tab:counting_visual_search}
\centering
\resizebox{\linewidth}{!}{\setlength{\tabcolsep}{6pt}
\renewcommand{\arraystretch}{1.0}
\vspace{-6mm}
\begin{tabular}{lccccccc}
\toprule
\multirow{2}{*}{\textbf{Model}} & \multicolumn{2}{c}{\textbf{Visual Counting}} & \multicolumn{3}{c}{\textbf{Visual Search}} & \multicolumn{2}{c}{\textbf{General}} \\
\cmidrule(lr){2-3} \cmidrule(lr){4-6} \cmidrule(lr){7-8}
& CountBench & PixmoCount & V* Bench & HR-Bench 4K & HR-Bench 8K & ChartQA & CharXiv \\
\midrule
\multicolumn{8}{c}{\textbf{Closed-Source MLLMs}} \\
\midrule
GPT-4o & 87.9 & - & 67.5 & 65.0 & 59.6 & 86.7 & 47.1 \\

\midrule
\multicolumn{8}{c}{\textbf{Open-Source MLLMs}} \\
\midrule

Llava-OneVision-7B & 82.3 & 54.4 & 72.7& 68.5 & 60.0 & 80.4 & 27.1 \\
Llava-OneVision-72B & - & 60.7 &  73.8 & 66.3 &  60.9 & 83.7 & - \\
InternVL2.5-8B & 55.9 & - & 73.7 & 72.0 & 65.5 & 82.8& 37.2 \\
InternVL3-8B & 80.3 & - & 70.2 & 70.5 & 70.0 & 86.1 & 38.3 \\
InternVL3-78B & - & - & 76.4 & 75.5 &  67.3 & 89.7 & 46.0 \\
Qwen2.5-VL-72B & 93.6 & 62.3 & 84.8 & 79.4 & 76.3 &  89.5 & 49.7 \\
Qwen2.5-VL-32B & 87.8 & 56.0 &  85.9 & 74.8 &  71.6 & - & 47.6 \\
Qwen2.5-VL-7B & 76.5 & 50.4 & 76.4 & 69.0 & 66.0 & 86.3 & 42.1 \\
\midrule
\multicolumn{8}{c}{\textbf{Tool-Integrated Open-Source MLLM}} \\
\midrule
Pixel Reasoner-7B & - & - & 84.3 &  72.9 & 66.9 &  -  & - \\
Deepeyes-7B$^{\dagger}$ & 80.4 & 57.2 & 90.4 & 74.8 & 71.9 & 78.2 & - \\
Thyme-VL-7B & 84.8$^{\dagger}$ & - & 82.2 & 77.0 & 72.0 & 86.1 & 44.2$^{\dagger}$ \\
\midrule
\rowcolor{darkgray}
CodeDance-7B & 91.2 & 77.1 & 84.8 & 75.2 & 72.3 & 87.5 & 44.1 \\
\multicolumn{1}{r}{\textit{$\Delta$ vs. Qwen2.5-VL-7B}} & \textcolor{green!60!black}{$\uparrow$\textbf{19.2}\%} & \textcolor{green!60!black}{$\uparrow$\textbf{53.0}\%} & \textcolor{green!60!black}{$\uparrow$\textbf{11.0}\%} & \textcolor{green!60!black}{$\uparrow$\textbf{9.0}\%} & \textcolor{green!60!black}{$\uparrow$\textbf{9.5}\%} & \textcolor{green!60!black}{$\uparrow$\textbf{1.4}\%} & \textcolor{green!60!black}{$\uparrow$\textbf{4.7}\%} \\
\bottomrule
\end{tabular}
}
\end{table*}

\textbf{Research gaps.} 
While recent models have made notable progress, fundamental gaps remain unresolved. 
(1) Current approaches largely extend CoT into multimodal reasoning via text-only templates, failing to incorporate new observations, refine intermediate steps, or validate its reasoning against visual evidence~\cite{ko2025flex,feng2025retool}. 
(2) In addition, o3 remains a proprietary black-box system: its internal mechanisms are inaccessible, its reasoning process is less transparent, and its outputs cannot be systematically studied or reproduced.
(3) Most open-source systems incorporating visual reasoning remain restricted to predefined visual workflows, or rigid and schema-based pipelines (e.g., predicting bounding box coordinates for cropping operations), which are inherently inflexible and task-specific, limiting transfer to new tools and tasks~\cite{zheng2025deepeyes,su2025pixel,zhang2025chain,su2025openthinkimg}.
(4) Existing methods~\cite{zhang2025thyme, zheng2025deepeyes} do not consider when a model should invoke tools, leading to tool underuse or overuse, as shown in \Cref{fig:teaser}.

Consequently, the field still lacks an open and verifiable medium that is general across tools and tasks for multimodal reasoning, allowing MLLMs to dynamically compose tools, produce intermediate artifacts, and self-check their outputs in a transparent and reproducible manner. Addressing this gap is crucial for achieving task-adaptive, explainable, and transferable reasoning across complex real-world tasks.


In this work, we introduce \textbf{\name}, a multimodal reasoning framework that leverages executable code as a unified medium for visual reasoning. Unlike prior schema-based pipelines with fixed operation templates, code enables the model to define, compose, and execute diverse visual–symbolic operations, producing both intermediate artifacts (e.g., cropped regions, plots, annotations) and final answers within a unified, verifiable reasoning process.
To equip the model with fundamental skills, we curate a high-quality trajectory dataset and use supervised fine-tuning (SFT) to teach atomic capabilities such as counting, spatial grounding, and image annotation, enabling an iterative exploration-reflection reasoning process. 
Building on this foundation, we employ reinforcement learning (RL) to further enhance tool-based reasoning. A central challenge we identify is a \textit{trade-off between exploration and selectivity}: naïve policies often overuse tools, incurring unnecessary steps, or underuse them, failing to leverage visual interactions when needed. To address this, we design a difficulty-adaptive tool-reward mechanism that adjusts incentives based on task complexity, encouraging longer operation chains for hard problems while discouraging unnecessary calls for simple ones. This aligns learning with the structure of multimodal tasks, resulting in more adaptive and transparent reasoning. Together, these components enable \name to advance beyond rigid schema-based methods and offer an open, generalizable medium for executable visual reasoning.

\textbf{Empirical observation of emergent behaviors.}
Although trained only on atomic operations, the framework exhibits behaviors beyond direct supervision during RL, including composing operations (e.g., localization followed by counting) and generating new procedural code for visual analysis. These behaviors arise from the base model’s pretrained knowledge and are driven by adaptive rewards rather than explicit instruction, highlighting the scalability of dynamic invocation for more flexible and general reasoning.

\textbf{Our contributions are summarized as follows}:  
\textbf{(1)} 
We introduce \name, a multimodal agent that can ``think with images'' by planning and composing visual–symbolic operations through executable code as a unified medium. To this end, we curate a 34K high-quality SFT dataset covering diverse atomic code capabilities (e.g., cropping, drawing, point plotting), and additionally design a difficulty-adaptive reward mechanism for RL, enabling multi-turn reasoning and balanced tool use.  
\textbf{(2)} 
We evaluate \name on more than 10 multimodal benchmarks, spanning both general perception and complex reasoning (e.g., visual search, math reasoning). Across multiple benchmarks, it outperforms advanced closed models (e.g., GPT-4o) and larger open-source baselines (e.g., Qwen2.5-VL-32B), demonstrating strong perception and reasoning capabilities, and broad generalizability.
\textbf{(3)} 
Despite being trained only on atomic operations, \name empirically exhibits emergent behaviors during RL training, including spontaneous novel tool routines, unseen operation compositions, and cross-task transfer to novel tasks. 
These promising observations highlight the scalability and generality of our framework, and we empirically verify them in ablation studies.

%% file: sec/2_RelatedWorks.tex
\section{Related Works}
\label{sec:relatedworks}

\textbf{Multimodal reasoning and tool invocation.}
Building upon text-based chain-of-thought (CoT) reasoning~\cite{wei2022chain,yao2022react}, researchers have extended intermediate reasoning steps to multimodal settings~\cite{zheng2025deepeyes,yeo2025demystifying} including counting~\cite{zhang2024counting}, localization~\cite{wu2024v}, charts~\cite{li2024graphotter}, and visual math~\cite{chen2025mint}.
To enhance reasoning capabilities, recent works integrate external tools through reasoning-and-acting frameworks~\cite{yao2022react,yang2023mm}, learned API usage~\cite{schick2023toolformer}, and multimodal agents that orchestrate OCR, detection, and editors~\cite{wu2023visual,shen2023hugginggpt}. ViperGPT~\cite{suris2023vipergpt} compiles queries into executable programs.
Recent models like OpenAI's o3~\cite{openai2025o3} integrate comprehensive tool capabilities directly into reasoning chains, trained via RL on large-scale CoT data. 
Other approaches include RL-based tool invocation~\cite{zheng2025deepeyes,su2025pixel} and SFT-based methods~\cite{wang2025simple}. 
However, challenges remain such as ad-hoc operations, sparse supervision, limited task coverage and lack of comprehensive evaluation. 
We diverge from prior work by pursuing code as a general medium to execute multimodal reasoning across diverse atomic abilities.

\noindent \textbf{Adaptive thinking capability of MLLMs.} 
%
Recent work on adaptive reasoning in LLMs explores how models can dynamically decide when and how much to think~\cite{zhang2025adaptthink} instead of relying on fixed chain-of-thought steps. Approaches such as adaptive thinking~\cite{wan2025adapthink,wu2025arm}, concise reasoning~\cite{shrivastava2025sample}, and learn-to-switch frameworks~\cite{zhang2025continue} show that LLMs can selectively perform deeper reasoning only when necessary, improving efficiency while preserving accuracy. Meanwhile, MLLMs can extend this: recent fast–slow vision-language reasoning demonstrates that models can also adjust reasoning depth based on visual–textual difficulty~\cite{xiao2025fastmllm}. However, these methods focus mainly on adjusting internal reasoning depth rather than deciding when visual operations or tools should be invoked. This limitation motivates the investigation of adaptive tool calling for multimodal reasoning, where MLLMs must think with visual inputs effectively.

%% file: sec/3_method.tex
\section{Methodology}
\label{sec:method}

\subsection{Overview}

\textbf{Think-execute-feedback as reasoning unit.}
An overview of our framework is shown in \Cref{fig:method}. 
Given a multimodal user query (including a text prompt and an image), the policy model (an MLLM) produces rollouts interleaving natural-language reasoning with executable code (e.g., cropping, plotting). Code is executed in a separate sandbox, and the resulting visual evidence (e.g., a cropped region) is concatenated with text to refine reasoning in the next turn or yield the final answer (e.g., ``blue and yellow'' in \Cref{fig:method}). 

We define a \emph{think–execute–feedback} cycle as the minimal reasoning unit under a policy model~$\pi$, where each turn comprises (i) the current query and reasoning trace, (ii) a candidate action, and (iii) the resulting observation after code execution. Formally, a trajectory is:
\[
\tau = \big( (s_{t_1}, a_{t_1}, s'_{t_1}), \ldots, (s_{t_{M-1}}, a_{t_{M-1}}, s'_{t_{M-1}}), (s_{t_M}, a_{\text{ans}}) \big),
\]
{\text{where $t$ is the time step}}.
\(s_t = (x, o_t, \epsilon_t)\) contains the original query \(x\), the accumulated reasoning trace \(o_t\), and interpreter feedback \(\epsilon_t\). Actions \(a_t\) are drawn from a space including tool calls (code snippets) and a terminal answer; executing code yields an observation and updates the state to \(s'_t\). By iterating \(a_t \sim \pi(\cdot \mid s_t)\) until a final answer is produced or a maximum turn budget \(M\) is reached, each turn becomes an executable and verifiable reasoning unit. 
Building on this formulation, Section~\ref{subsec:syndata} details the curation of a high-quality trajectory dataset encompassing diverse atomic abilities. This dataset provides the foundation for initializing the policy model through SFT, before advancing to RL.

\begin{table}[t]
\caption{Comprehensive results across math reasoning benchmarks. \textsuperscript{$\dagger$} denotes results from their official papers.}
\vspace{-3mm}
\label{tab:mm_reasoning}
\centering
\resizebox{\linewidth}{!}{
\setlength{\tabcolsep}{2pt}
\renewcommand{\arraystretch}{1.0}
\begin{tabular}{lcccc}
\toprule
\multirow{2}{*}{\textbf{Model}} & \multicolumn{4}{c}{\textbf{Math-Benchmark}} \\
\cmidrule(lr){2-5}
& MathVision & MathVista & MathVerse & WeMath \\
\midrule
\multicolumn{5}{c}{\textbf{Larger MLLMs without Reasoning}} \\
\midrule

GPT-4o & 36.5 & 63.4 & 35.3 & 44.2 \\

Qwen2.5-VL-72B & 38.1 & 74.8 & 57.6 & - \\
\midrule
\multicolumn{5}{c}{\textbf{Open-source Reasoning MLLMs}} \\
\midrule
R1-Onevision-7B $^{\dagger}$ & 29.9 & 64.1 & 40.0 & - \\
R1-VL-7B$^{\dagger}$ & 24.7 & 63.5 & 40.0 & - \\

\midrule
\multicolumn{5}{c}{\textbf{Open-source General MLLMs}} \\
\midrule
InternVL2.5-8B & 22.0 & 64.4 & 39.5 & 23.9 \\
Llava-OV-7B & 18.4 & 63.2 & 26.2 & 17.3 \\
Qwen2.5-VL-7B & 25.0 & 68.1 & 45.1 & 35.4 \\
DeepEyes-7B & 26.6 & 70.1 & 47.3 & 38.9 \\
\rowcolor{darkgray}
CodeDance-7B (Ours) & 29.6 & 70.3 & 46.8 & 39.6 \\
\bottomrule
\end{tabular}
}
\vspace{-6mm}
\end{table}
\textbf{Policy optimization via verifiable evidence.}
In the RL stage, we require a policy optimization method that can compare multiple rollouts and update the model accordingly. 
In our case, Group Relative Policy Optimization (GRPO)~\cite{shao2024deepseekmath} provides a natural baseline, as it directly normalizes rewards across sampled trajectories without relying on a separate value network. 
However, standard GRPO assigns a uniform advantage to all tokens within a trajectory, which limits its effectiveness for multi-turn tool reasoning requiring intermediate correction. 
To address this, we extend the reward design with sequence-level and turn-level components. 
In particular, each rollout is evaluated with a composite reward $r$ that integrates outcome and tool-related signals (also see \Cref{fig:method}, right side):
\[
r(\tau) = R_{\text{acc}}(\tau) + R_{\text{format}}(\tau) + R_{\text{\rname}}(\tau),
\]
where $R_{\text{acc}}$ denotes final-answer correctness and $R_{\text{format}}$ enforces format compliance, respectively.
In our design, we introduce a two-level reward $R_{\text{\rname}}$ (\textbf{Reward for Balanced Adaptive Tool Calling}) that advocates adaptive tool calling based on the task difficulty. 
In particular, it decomposes into a sequence-level $R_{\text{seq}}$ and a turn-level $R_{\text{turn}}$, balancing task difficulty with step-wise tool calling correctness.
Subsequently, the advantage $A$ is written as:
\[
A(\tau) = A_{\text{seq}}\!\big(R_{\text{acc}}, R_{\text{format}}, R_{\text{seq}}\big) + A_{\text{turn}}\!\big(R_{\text{turn}}\big).
\]
Next, we show that this formulation combines global, step-level trajectory outcomes with local, turn-level execution feedback, producing more adaptive tool invocation. 


\textbf{Sequence-level adaptive reward for code-invocation}.
Simply rewarding every successful tool call can  lead to degenerate behaviors such as tool spamming or reward hacking on trivial problems (see Appendix \Cref{fig:reward_hacking}), which may hinder the reasoning performance (ablation studies in \Cref{tab:ablation_reward}). 
To address this, we design an adaptive reward that conditions tool incentives on the group-level accuracy $\mu_{\text{acc}}$: when most rollouts already solve the task correctly (indicating that the problem is relatively easy or solvable without additional tool assistance), further invocations are discouraged. 
In contrast, low $\mu_{\text{acc}}$ encourages additional exploration.
Formally, the sequence-level reward is as follows:
\begin{equation}
\begin{aligned}
R_{\text{seq}} &=
\Big( 0.5 + 0.5 \cdot \mathbb{I}_{R_{\text{acc}}(\tau)>0} \Big)
\cdot d \cdot \frac{N_{\text{succ}}(\tau)}{N_{\text{total}}(\tau)},\\
d &= \sigma\bigl(\gamma(0.5 - \mu_{\text{acc}})\bigr) - \delta, \sigma(z) = \frac{1}{1 + e^{-z}},
\end{aligned}
\end{equation}
where $N_{\text{succ}}(\tau)$ and $N_{\text{total}}(\tau)$ denote the numbers of successful and total tool calls in trajectory $\tau$, and $d$ is a group-accuracy–dependent scaling factor that suppresses rewards for unnecessary tool calls on easy queries (high $\mu_{\text{acc}}$) and amplifies rewards for tool-assisted solutions on hard queries (low $\mu_{\text{acc}}$), thereby enabling adaptive tool invocation. 
Here, $\gamma$ and $\delta$ are hyper-parameters controlling the strength: higher $\mu_{\text{acc}}$ reduces $d$ (discouraging redundant calls), while lower $\mu_{\text{acc}}$ increases $d$ (promoting exploration). 
We set $\gamma=4$ and $\delta=0.2$ in our main experiments. 
See our Appendix \ref{appendix:hyperparams} for discussion of these hyper-parameters.

\textbf{Turn-level execution reward.}
To penalize failed executions and provide dense correction signals, we introduce a turn-level reward. 
For each turn $m$, an immediate penalty $r_{\text{turn},m} = -0.5$ is assigned if the code execution fails, and $0$ otherwise.  
To capture long-term effects, we recursively redefine $r_{\text{turn},m}$ as the accumulated discounted return:
{
\abovedisplayskip=4pt \belowdisplayskip=4pt
\begin{equation}
\begin{aligned}
R_{\text{turn}}^{m} =r_{\text{turn}}^{m}+\beta\cdot R_{\text{turn}}^{m+1},
\quad
A_{\text{turn}}^{m} = \frac{R_{\text{turn}}^{m} - \mu_{\text{batch}}}{\sigma_{\text{batch}}}.
\end{aligned}
\end{equation}
}
Here, $R_{\text{turn}}^{m}$ denotes the discounted cumulative turn-level return, 
$\beta$ is a discount factor, and $\mu_{\text{batch}}, \sigma_{\text{batch}}$ denote the batch-wise mean and standard deviation of $R_{\text{turn}}$, respectively.
This turn-level reward discourages credit assignment to incorrect intermediate reasoning steps, even when the final answer is correct, and it helps mitigate entropy collapse during training~(\Cref{fig:entropy}). The final advantage is obtained by combining the resulting $A_{\text{turn}}$ with the sequence-level advantage $A_{\text{seq}}$ (from outcome-level rewards, see Appendix). 

Together, the group-adaptive $R_{\text{seq}}$ evaluates the quality of an \emph{entire} trajectory, while $R_{\text{turn}}$ assesses the correctness of \emph{individual} tool calls. This complementary design, which we term $R_{\text{\rname}}=R_{\text{seq}}+R_{\text{turn}}$, mitigates reward hacking, balances efficiency with necessary exploration, and yields more robust multimodal reasoning policies. We discuss more design choices and results in ~\Cref{fig:entropy}.

\begin{figure*}[t] 
\newcommand{\colwid}{0.32\linewidth} 
    \centering 
    \vspace{-5mm}
    \includegraphics[width=0.92\linewidth]{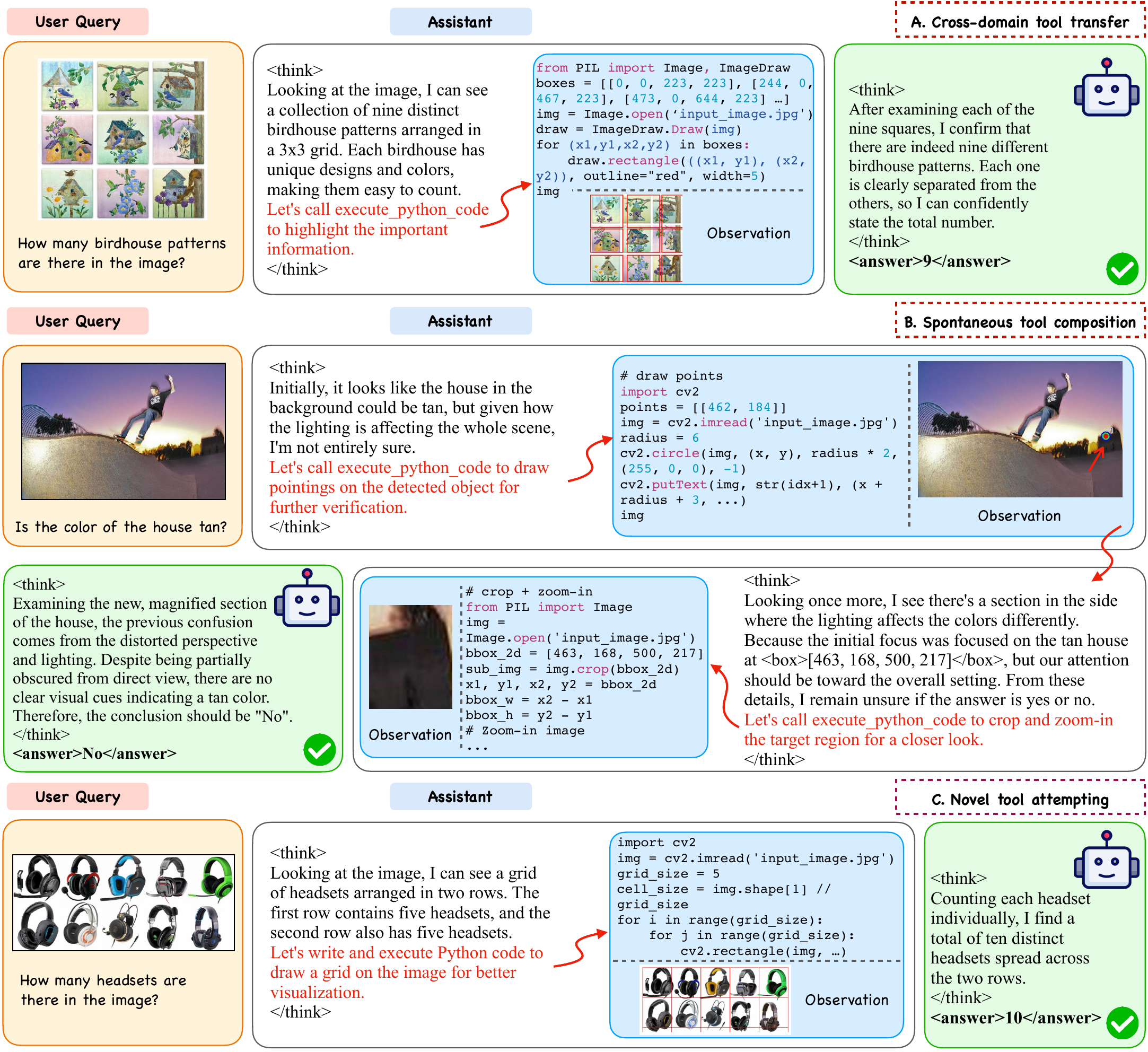} 
    \caption{
\textbf{Intriguing reasoning trajectories emerge during RL.}
These behaviors are absent from the SFT data and arise from pretrained knowledge further shaped by RL, reflecting our design of adaptive tool invocation. These emergent patterns motivate our scaling study in \Cref{fig:scaling}, where we further examine whether these capabilities strengthen with larger data, longer training, and bigger models.
}
    \label{fig:emergents} 
\end{figure*}
\subsection{Data Engineering}
\label{subsec:syndata}

In our design, the lack of high-quality, multi-turn, multimodal reasoning datasets that advocate adaptive tool calling with interleaved code trajectories remains a major barrier for training. 
To address this, we construct a 34k dataset for cold start, featuring \textbf{(1)} the reasoning trajectories that include code execution results appended as \textbf{visual evidence} in training, and \textbf{(2) adaptive tool invocation} based on the level of questions. 
For easy questions, the model is encouraged to answer directly without any reasoning process.
For hard questions, we enable code-based tool integration for executable reasoning.
Practically, to mitigate the issue of tool spamming, we ensure that the pipeline follows a two-step design: 
\textbf{(i) weak-to-strong filtering}, where public resources (e.g., SA1B, GEOqa\_plus, MMK12) are automatically filtered and stratified by difficulty using Qwen2.5-VL-7B; and 
\textbf{(ii) multi-turn atomic supervision}, where hard cases are decomposed into trajectories covering three categories: fundamental image transforms (crop, resize), mathematical computation (measurement, algebra, aggregation), and open-ended visual editing (drawing, annotation, etc.).
Each trajectory is further verified by a stronger MLLM (e.g., Qwen2.5-VL-72B) to cross-validate the correctness. 
The query, code, and response are embedded as follows:

\begin{tcolorbox}[colback=gray!5!white, colframe=gray!65!black, title=Multimodal CoT trajectories for cold-start, sharp corners=south]
\footnotesize
\noindent
\texttt{\textbf{Query}: $<$IMAGE$>$ Is the flag blue and yellow or red and yellow? \\
\textbf{Response}: The image shows..., Let's call execute\_python\_code: $\backslash$n \\<{\textcolor{Rubinered}{code}}>from PIL import Image $\backslash$n img = Image.open(‘img.jpg')...<{\textcolor{Rubinered}{/code}}>.\\
Appending compiling results... $\backslash$n <answer>blue and yellow</answer>
}\\
\texttt{\textbf{Example atomic abilities} (<{\textcolor{Rubinered}{code}}><{\textcolor{Rubinered}{/code}}>):}
\includegraphics[width=\linewidth]{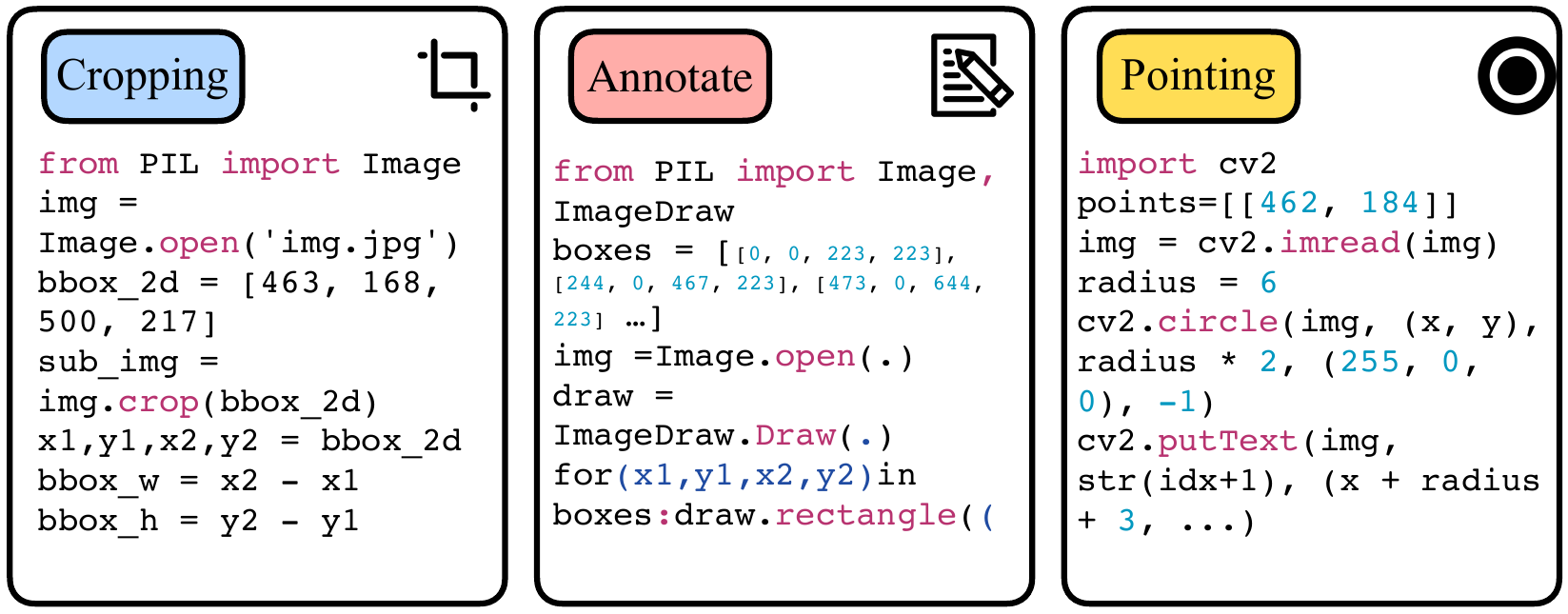}
\end{tcolorbox}

These trajectories provide verifiable supervision of the atomic skills used for SFT as cold-start. 
In Appendix~\ref{appendix:data_curation}, we provide the comprehensive recipe on trajectory synthesis, filtering and code integration for SFT data, with additional examples for RL training.

%% file: sec/4_exps.tex
\section{Experiments}
\label{sec:experiment}

\definecolor{rowCyan}{RGB}{236,246,252}      
\definecolor{rowPink}{RGB}{249,240,245}      
\definecolor{rowGreen}{RGB}{238,246,236}     
\definecolor{rowGray}{RGB}{245,245,245}      
\definecolor{rowSoftGray}{RGB}{239,239,239}  

\textbf{Implementation Details.}
Our method is designed to be seamlessly integrated into popular MLLM architectures. In our experiments, we mainly build on Qwen2.5-VL-7B~\cite{Qwen-VL} as the base model, and compare against both open-source reasoning MLLMs (e.g., DeepEyes~\cite{zheng2025deepeyes}) and advanced closed models (e.g., GPT-4o) across four benchmark categories: math reasoning (MathVista~\cite{lu2023mathvista}, MathVision~\cite{wang2024measuring}, MathVerse~\cite{zhang2024mathverse}, WeMath~\cite{qiao2024we}, general reasoning (ChartQA~\cite{masry2022chartqa}), counting (PixmoCount~\cite{pixmo}, CountBench~\cite{paiss2023countclip}), and visual search (V* Bench~\cite{wu2024v}, HRBench~\cite{hrbench}). 
For training, we adopt SWIFT~\cite{zhao2024swiftascalablelightweightinfrastructure} for SFT and VeRL~\cite{sheng2024hybridflow} for RL. 
We empirically set $\gamma=4$, $\delta=0.2$, and $\beta=0.2$.
To ensure a fair comparison, we adopt VLMEvalKit~\cite{duan2024vlmevalkit} as the evaluation framework. The max-turn set to 10 for evaluation and 6 for training.

\begin{table*}[htbp]
\vspace{-0.5em}
\caption{Ablation study on reward design. We report accuracy, average turns, and execution success rate on different reasoning benchmarks.}
\vspace{-0.5em}
\label{tab:ablation_reward}
\centering
\resizebox{\linewidth}{!}{
\setlength{\tabcolsep}{8pt}
\renewcommand{\arraystretch}{1.2} 
\begin{NiceTabular}{lccccccccc}
\toprule
\textbf{Components}
& \makecell{\textbf{CountBench} \\ \scriptsize{Acc. / Turns}}
& \makecell{\textbf{Pixmo} \\ \scriptsize{Acc. / Turns}}
& \makecell{\textbf{V* Bench} \\ \scriptsize{Acc. / Turns}}
& \makecell{\textbf{HR 4K} \\ \scriptsize{Acc. / Turns}}
& \makecell{\textbf{HR 8K} \\ \scriptsize{Acc. / Turns}}
& \makecell{\textbf{MVsion} \\ \scriptsize{Acc. / Turns}}
& \makecell{\textbf{MVerse} \\ \scriptsize{Acc. / Turns}}
& \makecell{\textbf{Avg.} \\ \scriptsize{Acc. / Turns}}
& \textbf{Succ.} \\
\midrule
SFT Cold-Start (w/o RL)
& \makecell{85.3 \\ \scriptsize{1.27}}
& \makecell{66.9 \\ \scriptsize{1.39}}
& \makecell{82.7 \\ \scriptsize{2.01}}
& \makecell{72.1 \\ \scriptsize{1.17}}
& \makecell{67.1 \\ \scriptsize{1.09}}
& \makecell{23.0 \\ \scriptsize{2.84}}
& \makecell{41.4 \\ \scriptsize{2.19}}
& \makecell{62.6 \\ \scriptsize{1.71}}
& 0.96\\
\midrule
RL with $R_{\text{acc}}$+$R_{\text{format}}$
& \makecell{\underline{89.0} \\ \scriptsize{1.22}}
& \makecell{72.9 \\ \scriptsize{1.54}}
& \makecell{\underline{84.8} \\ \scriptsize{1.00}}
& \makecell{74.5 \\ \scriptsize{1.03}}
& \makecell{\underline{71.1} \\ \scriptsize{1.04}}
& \makecell{\underline{28.6} \\ \scriptsize{1.97}}
& \makecell{\underline{46.5} \\ \scriptsize{2.17}}
& \makecell{66.8 \\ \scriptsize{\underline{1.42}}}
& \underline{0.97} \\
+$R_{\text{DeepEyes}}$~\cite{zheng2025deepeyes}
& \makecell{83.1 \\ \scriptsize{2.35}}
& \makecell{70.6 \\ \scriptsize{2.25}}
& \makecell{\textbf{86.4} \\ \scriptsize{2.01}}
& \makecell{\textbf{75.8} \\ \scriptsize{2.07}}
& \makecell{70.3 \\ \scriptsize{2.12}}
& \makecell{26.6 \\ \scriptsize{2.65}}
& \makecell{46.3 \\ \scriptsize{2.37}}
& \makecell{65.6 \\ \scriptsize{2.26}}
& \underline{0.97} \\

\rowcolor{gray!10}
+$R_{\text{BAT}}$-w/o Turn-level Reward
& \makecell{87.4 \\ \scriptsize{1.01}}
& \makecell{\textbf{78.8} \\ \scriptsize{1.00}}
& \makecell{\textbf{86.4} \\ \scriptsize{1.12}}
& \makecell{75.1 \\ \scriptsize{1.07}}
& \makecell{70.1 \\ \scriptsize{1.08}}
& \makecell{\underline{28.6} \\ \scriptsize{1.97}}
& \makecell{46.3 \\ \scriptsize{2.41}}
& \makecell{\underline{67.5} \\ \scriptsize{\textbf{1.38}}}
& 0.91 \\
\rowcolor{gray!10}
+$R_{\text{BAT}}$ (\name)
& \makecell{\textbf{91.2} \\ \scriptsize{1.00}}
& \makecell{\underline{77.1} \\ \scriptsize{1.00}}
& \makecell{\underline{84.8} \\ \scriptsize{1.00}}
& \makecell{\underline{75.2} \\ \scriptsize{1.06}}
& \makecell{\textbf{72.3} \\ \scriptsize{1.06}}
& \makecell{\textbf{29.6} \\ \scriptsize{2.19}}
& \makecell{\textbf{46.8} \\ \scriptsize{2.35}}
& \makecell{\textbf{68.1} \\ \scriptsize{\textbf{1.38}}}
& \textbf{0.99} \\
\bottomrule 
\end{NiceTabular}}
\end{table*}

\begin{figure*}[t]
    \includegraphics[width=\linewidth]{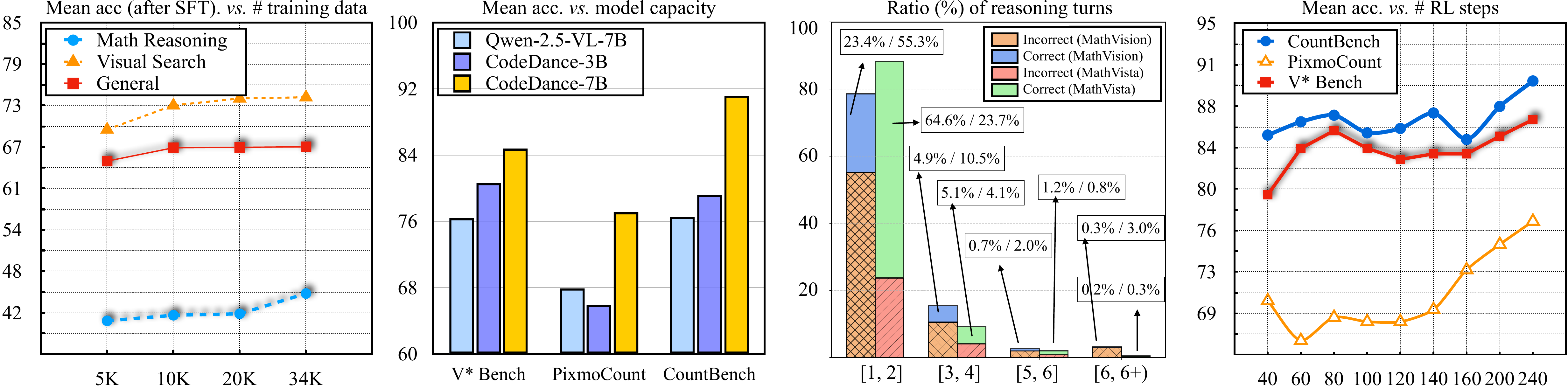}
    \caption{
    Scaling up compute budget on four dimensions: dataset size for SFT, model capacity, max-turns during inference and RL steps.
    }
\label{fig:scaling}
\end{figure*}

\subsection{Visual Reasoning Tasks}
\label{subsec:visual_centric}

As shown in Table~\ref{tab:counting_visual_search}, \name attains strong results across diverse visual reasoning benchmarks, including counting, visual search, and chart understanding. 
Notably, it achieves state-of-the-art performance on Counting and ChartQA, outperforming the baseline by a large margin and surpassing even larger models.
These improvements highlight the advantage of executable code as a reasoning medium: by delegating fine-grained visual analysis to code-based tools, it extends beyond the raw perceptual capacity of the base model, yielding gains that cannot be achieved through scaling alone, particularly on perception-heavy tasks.

\subsection{Math Reasoning Tasks}
\label{subsec:math_reasoning}
In mathematical reasoning, \name shows consistent gains over open-source baselines (see Table~\ref{tab:mm_reasoning}). 
For instance, it improves accuracy on MathVision from 25.0 to 29.6 (+18.4\%) and on WeMath from 35.4 to 39.6 (+11.9\%), while maintaining competitive results on other benchmarks. 
These tasks require precise symbolic manipulation and stepwise calculations, which are naturally supported by executable code. By externalizing intermediate steps into verifiable scripts, \name demonstrate strong accuracy and reliability than relying solely on internal approximation.

\subsection{Key Findings: Emergent Behaviors during RL}

Throughout the RL process, we observe novel and surprising empirical findings (shown in \Cref{fig:emergents}) that go beyond the atomic supervision provided during SFT. 
These findings point toward the scalability of code as a general reasoning medium, and we empirically study the potential in \Cref{fig:scaling}.

\textbf{Cross-domain tool transfer.}
We observe an emergent generalization ability in our \name, 
where visual operations defined for a specific task can be repurposed in other contexts. 
For example, the bounding-box operation was initially designed to highlight particular results within chart tasks in our SFT data. 
However, the model demonstrates the ability to adapt this operation for counting tasks during RL training: e.g. In Figure \ref{fig:emergents}A, the MLLM assistant first localizes all candidate objects by drawing bounding boxes, then validates the correctness of each localization, and subsequently derives the final count. 
Additional tool transfer trajectories can be found in \Cref{fig:emergents_supp1}.
Such behavior indicates that task-specific visual operations are not rigidly bound to their original purpose, 
but can be flexibly generalized to support broader multimodal reasoning scenarios. 
This suggests that visual operations such as bounding boxes can function as general \emph{reasoning primitives}, 
serving as transferable building blocks across heterogeneous tasks.

\textbf{Novel tool composition of learnt capabilities.}
Although during SFT data curation and collection, each task was restricted to a single predefined tool or coding operation, 
we observe that after post-training the model develops the ability to compose multiple atomic operations to address more complex tasks beyond the training coverage:
In \Cref{fig:emergents}B, to validate the color of the house, the MLLM assistant first applies a pointing operation to check the house position correctness, then use crop with zoom-in to focus on fine-grained details.
Similarly, we show example in Appendix \Cref{fig:emergents_supp2}
that the bounding-box drawing and crops are combined to focus and solve the chart reasoning task. 
These observations highlight the emergence of novel tool compositions, 
where elementary visual operations are flexibly combined to form higher-level reasoning strategies. 

\begin{figure}[t]
    \centering
    \includegraphics[width=0.88\linewidth]{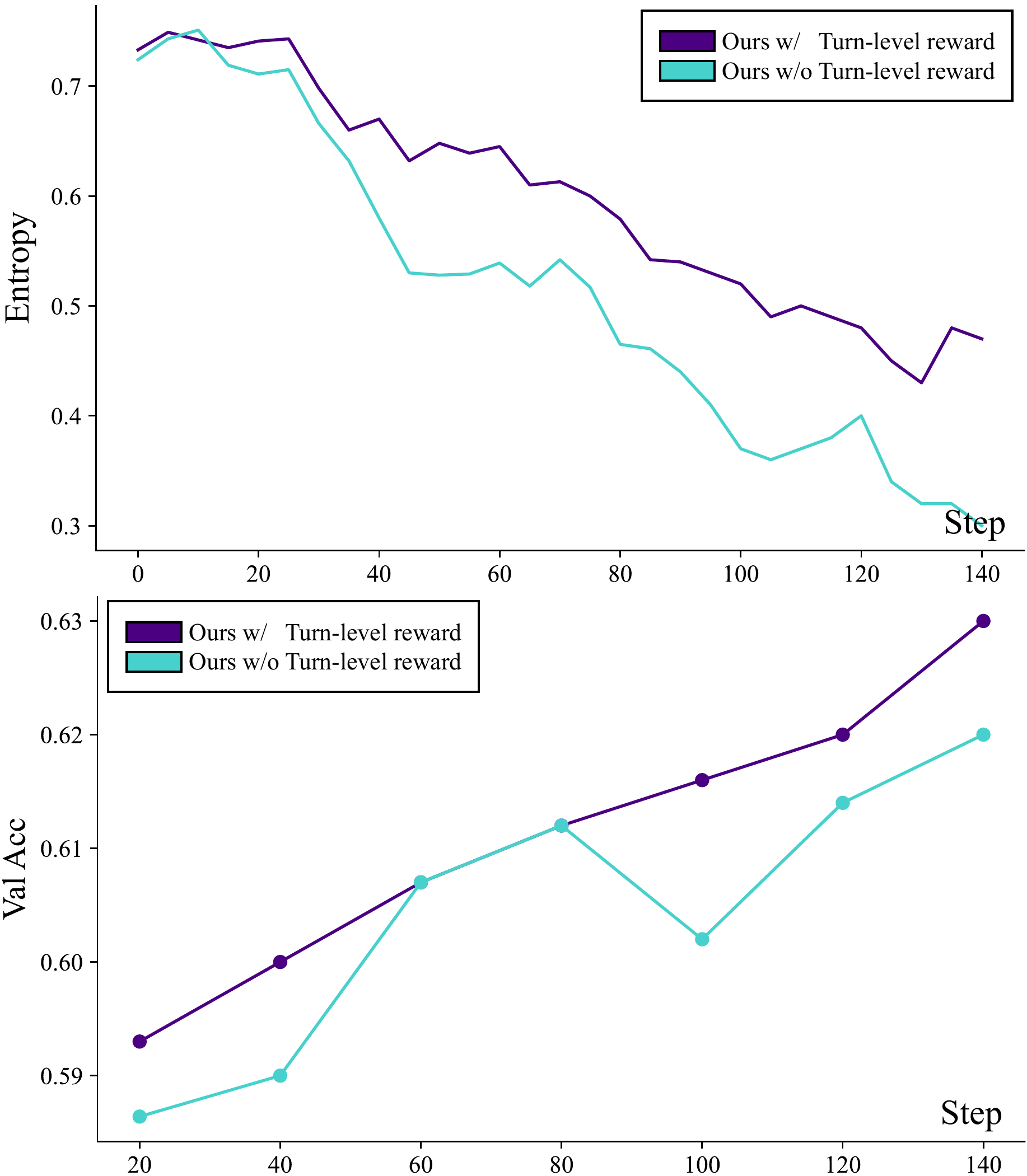}
    \vspace{-3mm}
    \caption{
    Entropy and validation accuracy of model's generation.
    }
    \vspace{-4mm}
    \label{fig:entropy}
\end{figure}

\textbf{Incentivizing emergence of novel unseen capabilities.}
Interestingly, we also find that the model exhibits a certain potential to generate tool codes not explicitly defined in the SFT data. These codes appear to be drawn from the model’s pretraining knowledge and are occasionally activated during the post-training stage. 
For example, when asked to count the number of headsets in an image (Figure \ref{fig:emergents}C), the MLLM does not directly respond with a number, but instead attempts to write Python code with OpenCV functions (e.g., using \texttt{cv2.rectangle} to overlay a grid for better visualization). 
This observation suggests that, beyond reproducing SFT-defined behaviors, the model attempts to reuse and adapt pretrained capabilities (e.g. complex OpenCV operations) to support reasoning tasks, indicating a certain potential for more flexible tool usage.

\subsection{Ablation Studies}
\label{subsec:ablation}


\textbf{Validation of reward design.}
We compare three reward designs for guiding tool usage (Table~\ref{tab:ablation_reward}):
(i) GRPO-reward (\emph{Outcome-level reward}) focuses only on final-answer correctness. While it shortens interaction turns, it discourages tool usage and underperforms on complex tasks.
(ii) \emph{DeepEyes reward} grants positive signals for every successful tool execution upon accurate answer. Although this encourages exploration, it also leads to tool overuse on trivial problems, increasing turns without consistent accuracy gains. A qualitative example is provided in Appendix \Cref{fig:reward_hacking}.
(iii) \emph{Our reward $R_{\text{\rname}}$} for adaptive tool-call balances the two extremes by penalizing redundant calls and rewarding selective, high-impact interactions.
As Table~\ref{tab:ablation_reward}, outcome-level GRPO under-utilizes tools, and the \textit{DeepEyes reward} inflates turns without reliable accuracy improvement. 
Furthermore, removing the turn-level reward severely degrades the execution success rate.
Overall, the full $R_{\text{\rname}}$ achieves the best overall accuracy while avoiding unnecessary tool use, consistently surpassing both baselines.

\textbf{Validation on scaling up the compute budget.}
Motivated by the emergent behaviors observed in Figure~\ref{fig:emergents}, we further examine whether \name continues to benefit from scaling along four axes, as summarized in Figure~\ref{fig:scaling}.
\textbf{(i) Training dataset size} for SFT, including Math Reasoning (MathVision, MathVerse, MathVista), Visual Search (V*, HR 4K, HR 8K), and General (ChartQA, MMVet, MMStar). We consistently observe that enlarging the SFT dataset from 5K to 34K yields steady accuracy gains, showing that both tool selection and symbolic planning benefit from broader coverage.
\textbf{(ii) Increasing model capacity} from 3B to 7B substantially boosts reasoning benchmarks such as counting and search, where CodeDance-3B even outperforms a stronger 7B model.
\textbf{(iii) Extending RL training} up to 240 steps further improves accuracy without overfitting, supported by our reward design $R_{\text{\rname}}$.
\textbf{(iv) Increase the inference turn budget}.
Although we set a maximum of 6 turns of the RL training, we observe that allowing more turns at inference (10) continues to improve reasoning performance. 
In experiments, the model achieves additional gains even beyond 6 turns ($0.3\%$), suggesting that the learned policy can generalize to longer reasoning horizons.

\textbf{Empirical study on inference entropy and accuracy.}
In \Cref{fig:entropy}, we evaluate the impact of incorporating turn-level reward ($R_{\text{turn}}$) on training dynamics and generalization.
We study the impact of turn-level reward in $R_{\text{\rname}}$ on several benchmarks, mainly including visual search (V$^*$ Bench), math reasoning (MathVista, MathVerse, MathVision), and counting benchmarks (PixmoCount and CountBench).
(a) \emph{Entropy:} 
Without $R_{\text{turn}}$, policy entropy collapses quickly because flawed intermediate steps may still lead to correct final answers, reinforcing shortcuts and limiting exploration~\cite{yu2025dapo}. With $R_{\text{turn}}$, intermediate penalties delay collapse, sustaining exploration.
(b) \emph{Validation Accuracy:} 
The additional corrective signals prevent premature convergence and translate into consistently higher accuracy, showing that local feedback improves global generalization.


%% file: camera_ready/sec/5_discussion.tex
\section{Discussion}
\label{sec:discussion}

We present \name, a task-aware, difficulty-adaptive MLLM that uses executable code for tool-integrated visual reasoning. 
By allowing models to define and execute code, \name enables flexible reasoning with interpretable intermediate evidence, while promoting adaptive tool calls and discouraging tool spamming. 
During RL, \name exhibits emergent behaviors beyond supervised skills, including novel tool routines, compositional strategies, and cross-domain transfer. 
We further verify its scalability, observing clear and consistent gains with increased compute budget.
Even at the 7B scale, \name achieves competitive performance across diverse benchmarks, highlighting code-based, difficulty-adaptive reasoning as a scalable and transferable paradigm for multimodal and agentic AI.


%% file: camera_ready/sec/X_suppl.tex
\section*{Appendix}
\label{sec:appendix}
This appendix serves to enhance the reproducibility and transparency of our work by supplementing the main paper with critical technical details and experimental analyses. We provide additional implementation details (e.g., pipeline and examples of the trajectory construction, training setups, and hyperparameters), further descriptions and comparisons of our curated datasets and prompt templates, additional qualitative and quantitative results on various benchmarks, and extended discussions on design choices, broader impacts, limitations and future works.
{
\tableofcontents
}

\section{Additional implementation details}
\label{appendix:implementation}

\subsection{Hyperparameters in training and inference}

In \Cref{tab:params}, we present the additional hyperparameters used for training our model on the multimodal reasoning tasks. 
We primarily adhere to the same settings as Qwen2.5-VL~\cite{Qwen2.5-VL}, and these parameters are mostly applied across other tasks.

\begin{table}[h]
    \centering
    \caption{Hyperparameters used in training/inference.} \label{tab:params}
    \setlength{\tabcolsep}{4pt}
    \renewcommand{\arraystretch}{1.1}
    \begin{tabular}{c|l|c}
        \toprule
                                   & Param. Name & Value / Type \\ \hline
        \multirow{3}{*}{SFT}  & Batch size                  & 128 \\
                                   & Learning rate & 5e-5 \\
                                   & Warmup ratio               &  0.05 \\ \hline
        \multirow{5}{*}{RL}       & Numerical precision   &  BF16 \\
                                   & Global batch size            &  256 \\
                                   & Rollout            &  8 \\
                                   & Total epochs          &  1 \\
                                   & Time                  &  $\sim$2 Days \\ \hline
        \multirow{1}{*}{Inference \& Eval} & Deployment platform   &  vLLM~\cite{kwon2023efficient} \\
        \bottomrule
    \end{tabular}
\end{table}

\begin{figure*}[t]
    \centering
    \includegraphics[width=1\linewidth]{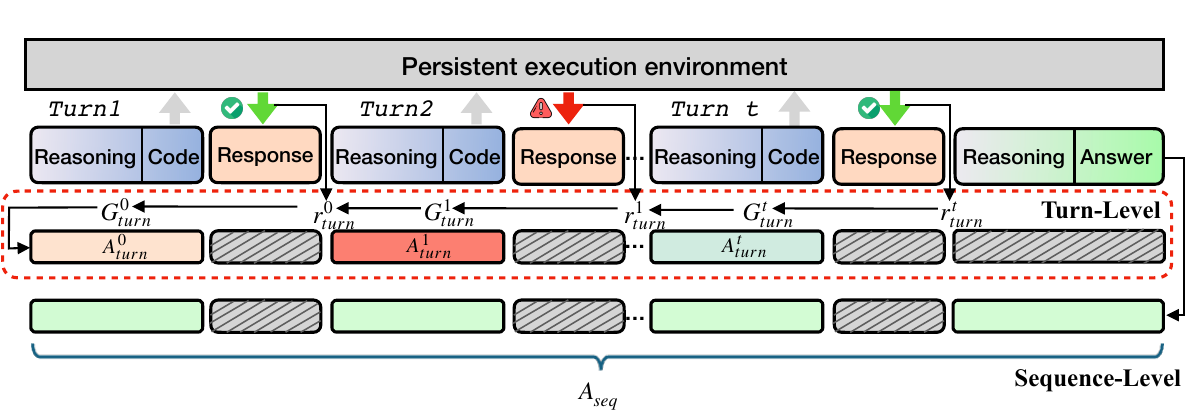}
    \caption{Interaction of our reward design $R_{\text{\rname}}$ with the code sandbox (i.e., the execution environment).}
    \label{fig:sandbox_app}
\end{figure*}

\subsection{Implementation of Code sandbox}
\label{appendix:sandbox}

To safely execute model-generated programs during training and evaluation, we run all Python code in an internal sandboxed environment. Each execution instance is assigned an isolated working directory and a dedicated namespace for variables, functions, and imports; for multimodal tasks, input images are stored in the corresponding temporary directory and are only visible within that instance. This design avoids interference across concurrent rollouts and confines state to a single trajectory.

Before running user code, the sandbox disables a small set of APIs that may compromise the stability (e.g., blocking user input, direct process termination), and treats each execution as an atomic transaction: if a run fails or is interrupted, partial updates are discarded and the logical state of the instance reverts to the last successful step. We further enforce a strict per-call wall-clock time limit (15s by default); when the limit is exceeded, the current execution is aborted and a timeout status is returned. 
Finally, the sandbox centralizes observability by capturing structured stdout/stderr logs and graphical outputs produced by common plotting libraries (e.g., \texttt{matplotlib}, \texttt{PIL}), which are then fed back to the model as textual and visual observations. 
An illustration of interaction with the code sandbox is shown in \Cref{fig:sandbox_app}.

\subsection{Implementation of standard GRPO}
\label{appendix:grpo}
Here, we reveal additional implementation details regarding the RL algorithm used in our work. 
Group Relative Policy Optimization (GRPO,~\citet{shao2024deepseekmath}) has demonstrated strong effectiveness across diverse tasks, particularly in multi-turn tool call agents and ``thinking with images'' system~\cite{feng2025retool,fu2025areal,zheng2025deepeyes,su2025pixel}. 
Unlike PPO~\cite{schulman2017proximal}, GRPO removes the need for a separate value network by directly computing advantages from the normalized rewards of $G$ sampled solutions. 
Formally, let $\pi_{\theta_{\text{old}}}$ and $\pi_{\theta}$ denote the policy model (parameterized by $\theta$) before and after the update, respectively, both defined over the action/token space at each position. 
For a question $q$ sampled from a task dataset $\mathcal{Q}$, a group of $G$ candidate solutions $\tau_i \sim \pi_{\theta_{\text{old}}}$ are rolled out and evaluated with a reward function $r(\cdot)$. 
Building on the clipped surrogate objective of PPO, we write the objective $\mathcal{J}$ in an empirical expectation form:
\begin{align}
\mathcal{J}_{\text{GRPO}}(\theta)
&= \mathbb{E}_{q\sim \mathcal{Q},\,\{\tau_i\}_{i=1}^G \sim \pi_{\theta_{\text{old}}}(\cdot\mid q)} \Bigg[
\frac{1}{G} \sum_{i=1}^G \frac{1}{|\tau_i|}\\
\sum_{t=1}^{|\tau_i|}
\min\Big(
r_{i,t} A_i, \nonumber
&
\text{clip}(r_{i,t}, 1-\varepsilon, 1+\varepsilon) A_i
\Big)
\Bigg].
\label{eq:grpo}
\end{align}
where
\(
r_{i,t}
= \frac{\pi_\theta(\tau_{i,t} \mid q, \tau_{i,<t})}
       {\pi_{\theta_{\text{old}}}(\tau_{i,t} \mid q, \tau_{i,<t})}
\), $\varepsilon=0.2$ by default, and $\text{clip}(\cdot)$ denotes the clipping operator for stability. 
We omit the KL penalty here for simplicity.
The normalized within-group reward then defines the advantage $A_i$ of solution $\tau_i$: 
\begin{gather}
    A_i = \frac{r(\tau_i) - \text{mean}(\{r(\tau_j)\}_{j=1}^G)}{\text{std}(\{r(\tau_j)\}_{j=1}^G)}.
\end{gather}
We mostly followed the original implementation of GRPO~\cite{shao2024deepseekmath} to compute outcome-driven advantage $A_{seq}$.


\section{Data engineering}

\subsection{Dataset curation}
\label{appendix:data_curation}
Here, we firstly discuss several related works on data synthesis for MLLM training to complete the related works. Then, we show details of our dataset curation pipeline.

\noindent
\textbf{Background: Synthetic reasoning data for MLLM post-training.}
High-performance MLLMs require substantial instruction-following training data with detailed reasoning trajectories.
Recent approaches include converting existing datasets using fixed templates~\cite{wei2021finetuned,dai2023instructblip} or distilling knowledge from strong teacher models~\cite{chen2024sharegpt4v,zhang2025thyme,wang2025simple}, with a focus on developing specific capabilities such as visual-centric reasoning~\cite{lan2024text4seg} and mathematical problem-solving assisted by visual cues~\cite{gao2023g,chen2025mint}.
However, several limitations persist in existing approaches:
(i) tool-grounded verification mechanisms are often absent, and
(ii) visual operations are typically limited to fixed schema such as cropping or zooming in~\cite{zheng2025deepeyes,su2025pixel}.
In contrast, we synthesize and curate training data with comprehensive reasoning trajectories and tool/code-assisted responses across a wide range of atomic visual operations, employing enhanced process supervision including multi-judge filtering and consistency validation. This leads to ``thinking with images'' reasoning capability~\cite{openai2025o3} with competitive performance while requiring substantially less training data.

\begin{figure*}[t] 
    \centering 
    \includegraphics[width=0.95\linewidth]{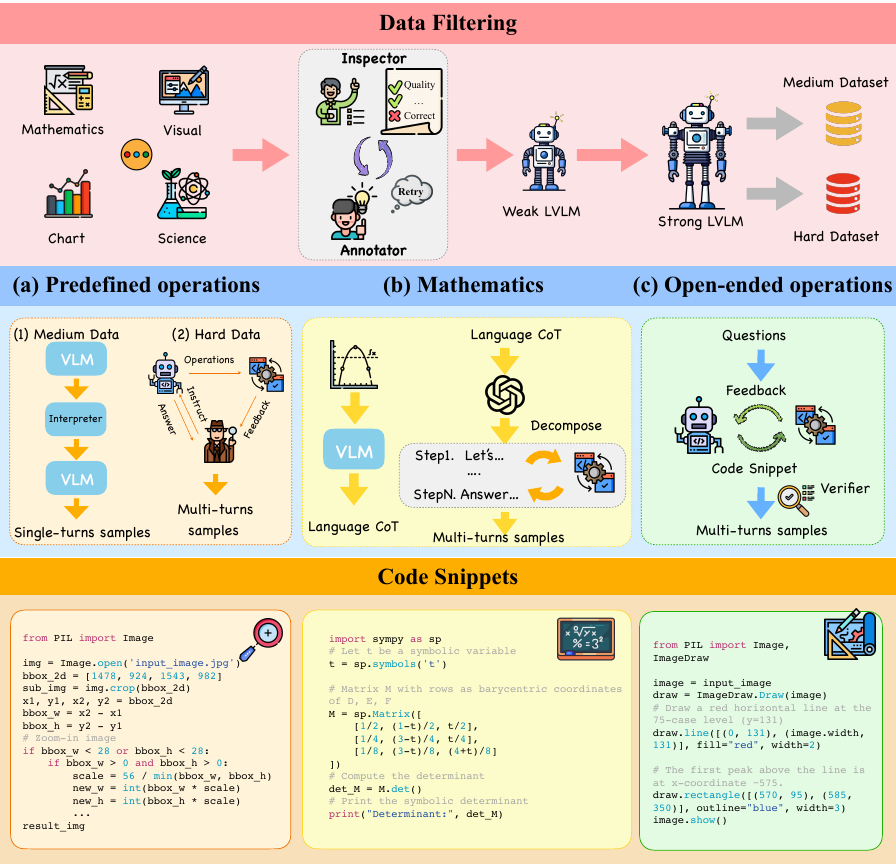} 
    \vspace{-3mm}
    \caption{\textbf{Overview of the SFT dataset curation pipeline.}
    \textbf{Top}: Weak-to-strong filtering. Candidate samples from diverse domains (mathematics, science, visual reasoning, charts) are validated for quality and correctness through automated inspector modules. A weak VLM removes trivial or low-information cases, while a stronger VLM further stratifies the remaining data into medium- and hard-difficulty sets.
    \textbf{Middle}: Multi-turn atomic supervision. The curated data are organized into three categories.
    (a) Predefined image operations (e.g., crop, resize, rotate), where medium samples produce single-turn trajectories and hard samples produce multi-turn reasoning with interpreter feedback.
    (b) Mathematical reasoning, where language-based CoT traces are decomposed into step-level atomic operations and converted into executable code.
    (c) Open-ended visual operations (e.g., drawing, annotation), where questions, feedback, and snippet verification form multi-turn executable trajectories.
    \textbf{Bottom}: Example executable snippets illustrating the supervision format across image processing, symbolic computation, and visual annotation.
    \textbf{A detailed curation pipeline example is shown in \Cref{fig:trajectory-syn-crop}.}
    }
    \label{fig:data-pipeline} 
\end{figure*}

\begin{figure*}[ht]
    \centering
    \vspace{10mm}
    \includegraphics[width=1\linewidth]{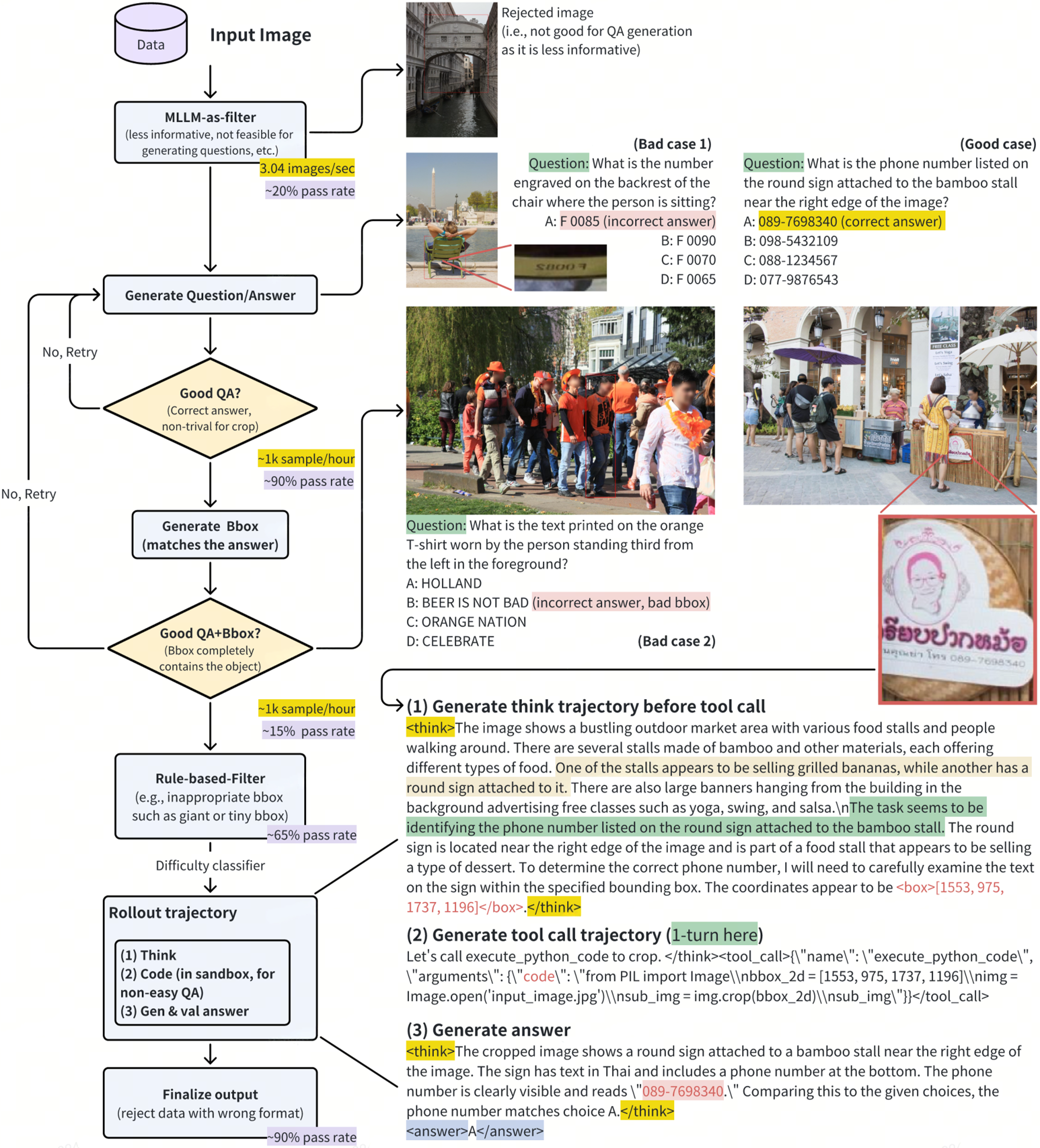}
    \caption{\textbf{An illustration of curating tool-integrated reasoning trajectories} (for cropping operations here). 
    Representative good and bad cases are shown on the right, covering incorrect answers, trivial or mismatched bounding boxes, and fully valid QA pairs. A rule-based filter further removes outputs with implausible bounding boxes.
    Similar operations are applied when synthesizing other trajectories with different atomic capabilities. Best viewed in color with zooming in.}
    \label{fig:trajectory-syn-crop}
\end{figure*}

\noindent
\textbf{Synthesize high-quality cold-start trajectories for tool-integrated reasoning (TIR).} An overview of the curation pipeline as a supplement of the main paper is shown in \Cref{fig:data-pipeline} and a detailed example is in \Cref{fig:trajectory-syn-crop}.
In addition, the dataset used for our trajectory synthesis is primarily composed of the following datasets:
\begin{itemize}
    \item \textbf{Mathematical Reasoning}: MMK12~\cite{meng2025mm}, Retool~\cite{feng2025retool}.
    \item \textbf{Table Data}: ChartQAPro~\cite{masry2025chartqapro}, chartgemma~\cite{masry2024chartgemma}.
    \item \textbf{Natural Images}: SA1B~\cite{kirillov2023segment}.
    \item \textbf{General Data}: Mulberry~\cite{yao2024mulberry}.
\end{itemize}
For the RL training, our data mainly comes from Deepeyes~\cite{zheng2025deepeyes},  SA1B~\cite{kirillov2023segment} and PixmoCount train~\cite{pixmo}.

\subsection{Prompt templates}
\label{appendix:prompt}

\noindent
\textbf{Prompt templates used in SFT data synthesis.}
To ensure the reliability and consistency of synthesized data, we design a set of standardized prompt templates tailored for different stages of the vision-language data pipeline. These templates serve complementary purposes: 
\textbf{(i) In \Cref{tab:image_informativeness_prompt}:} assessing the informativeness of candidate images to guarantee sufficient visual complexity for fine-grained reasoning;
\textbf{(ii) In \Cref{tab:detect_object_for_question}:} labeling and locating the objects that most match the question.
\textbf{(iii) In \Cref{tab:validation_prompt}:} validating the quality of automatically generated visual question–answer pairs;
\textbf{(iv) In \Cref{tab:step_answer_prompt}:} enforcing a structured step-by-step reasoning process with explicit final answers; and 
\textbf{(v) In \Cref{tab:revised_thinking_prompt}:} enhancing reasoning accuracy by incorporating code interpreter support for precise numerical or logical calculations. Together, these prompt templates provide a comprehensive and systematic framework for controlling data quality during synthesis, thereby improving the robustness and utility of the resulting multimodal datasets.

\begin{table}[h]
    \centering
    \caption{Prompt template for assessing image informativeness.}
    \label{tab:image_informativeness_prompt}
    \setlength{\tabcolsep}{6pt}
    \renewcommand{\arraystretch}{1}
    \begin{tabular}{p{0.95\linewidth}}
        \toprule
        You are an expert vision-language analyst. \\
        \textbf{Task} \\
        1. Observe the entire image. \\
        2. Decide whether the picture meets \textbf{all Four conditions} below: \\
        A. Diversity -- Contains $\geq$ 4 different object categories \textbf{or} $\geq$ 6 individual objects. \\
        B. Distinguishability -- Includes at least one object that is mostly un-occluded, covers < 30\% of the image area, and is not repeated by many visually identical copies. \\
        C. Zoom-in Benefit -- For that object (or another), some informative fine-grained detail (e.g., printed text, small logo, numerical value, subtle texture, or facial expression) would become noticeably clearer if the region were enlarged. In other words, a close-up view would materially help a downstream model answer a question about that object. \\
        D. Is it suitable to come up with some VQA questions that require fine-grained understanding? \\
        3. If \textbf{all} A, B, C, D are satisfied, Please respond with ``True'' or ``False''. \\
        \bottomrule
    \end{tabular}
\end{table}

\begin{table}
    \centering
    \caption{Prompt template for bbox generation.}
    \label{tab:detect_object_for_question}
    \setlength{\tabcolsep}{6pt}
    \renewcommand{\arraystretch}{1.2}
    \begin{tabular}{p{0.98\linewidth}}
        \toprule
        Please detect the entire object that most matches the question in the image. \\
        \textbf{Question:} \verb|{question}| \\
        If the target is part of an object, you need to give the bbox of the entire object. \\[2pt]\\
        For each object, return: \\
        - \verb|'label'|: the object name \\
        - \verb|'bbox_2d'|: the object's bounding box coordinates as \verb|[x1, y1, x2, y2]|. \\[2pt]\\
        Respond in a \textbf{JSON array}, where each entry is a dictionary with \verb|'label'| and \verb|'bbox_2d'|. \\
        \bottomrule
    \end{tabular}
\end{table}

\begin{table}[h]
    \centering
    \caption{Prompt template for visual question validation.}
    \label{tab:validation_prompt}
    \setlength{\tabcolsep}{6pt}
    \renewcommand{\arraystretch}{1.2}
    \begin{tabular}{p{0.98\linewidth}}
        \toprule
        You are a quality control assistant. Your task is to evaluate a visual question based on the provided image, question, and correct answer. \\
        \textbf{Image:} [Image is attached] \\
        \textbf{Question:} \verb|{question}| \\
        \textbf{Provided Correct Answer:} \verb|{correct_answer}| \\[2pt]\\
        \textbf{Evaluation Criteria:} \\
        1. \textbf{Correctness:} Is the provided ``Correct Answer'' truly the correct answer based on the image? \\
        2. \textbf{Difficulty:} Is the question non-trivial? It should require careful observation of details and not be something overly simple or obvious (e.g., ``What color is the sky?''). \\[2pt]\\
        \textbf{Your Response:} \\
        Respond with ``GOOD'' if the question meets BOTH criteria. \\
        Respond with ``BAD'' if the question fails one or both criteria. Do not provide any other explanation or text. \\
        \bottomrule
    \end{tabular}
\end{table}

\begin{table}[h]
    \centering
    \caption{Prompt template for step-by-step solving with answer tag.}
    \label{tab:step_answer_prompt}
    \setlength{\tabcolsep}{6pt}
    \renewcommand{\arraystretch}{1.2}
    \begin{tabular}{p{0.98\linewidth}}
        \toprule
        Solve the following problem step by step and then provide the final answer. \\
        The final answer MUST BE enclosed within \verb|<answer>| \verb|</answer>| tags. \\
        \textbf{Question:} \verb|{question}| \\
        \bottomrule
    \end{tabular}
\end{table}

\begin{table*}[h]
    \centering
    \caption{Prompt template for revised thinking with code interpreter.}
    \label{tab:revised_thinking_prompt}
    \setlength{\tabcolsep}{3pt}
    \renewcommand{\arraystretch}{1.15}
    \begin{tabular}{p{0.95\linewidth}}
        \toprule
        You are a helpful AI assistant. Initially, when solving a question, you would need to think step by step, without the ability to use code for calculation. Now, you have the capability to write code to use the code interpreter for calculation. The code will be executed by a sandbox, and the result can be returned to enhance your reasoning process. You can now leverage code to enhance your calculation while still maintaining the reasoning process. \\
        The thinking process can have multiple code snippets. Each code snippet is wrapped with \\
        \verb|<code>|\\
        \verb|```python|\\
        \verb|code snippet|\\
        \verb|```|\\
        \verb|</code>|\\[2pt]\\
        The returned result is wrapped with \\
        \verb|<interpreter> execution results</interpreter>|\\[2pt]\\
        \textbf{Goal:} Modify the original thinking process to make it more accurate by replacing manual calculation steps that can benefit from code execution with the corresponding code snippets and their interpreter's execution results. The core reasoning logic from the original thinking process, including any unsuccessful attempts, should remain unchanged. You should only replace the necessary manual calculation steps with code and interpreter's execution results, without altering the rest tokens of the thinking process. \\
        \textbf{User Question:} \verb|{question}| \\
        \textbf{Original Thinking Process (without code interpreter's support):} \\
        \verb|<original_thinking_process>| \verb|{original_response}| \verb|</original_thinking_process>| \\[2pt]\\
        \textbf{Details:} \\
        1. Identify sections where code execution could speed up the reasoning process or make the calculation more accurate. For simple calculations, you should keep the original text-based reasoning process without executing any code. \\
        2. Replace the manual calculation steps with code snippets and the corresponding interpreter's execution results. \\
        3. Keep the logical flow of the reasoning process intact, including any failed exploration attempts that were part of the initial process. \\
        4. The code snippets should be complete scripts, including necessary imports. \\
        5. Outputs in the code snippets must explicitly call the \textbf{print} function. \\
        6. Execution results should match the model's output exactly, with no extra or missing tokens. \\
        7. If, during the revised thinking process, you obtain the same result as in the original reasoning, you may omit numerical computations and refrain from simplifying to specific numeric values. \\
        8. If the Original Thinking Process does not include an \verb|<answer>| section at the end, please add it: \verb|<answer> \boxed{{'The final answer goes here.'}} </answer>| \\[4pt]\\
        \textbf{Revised Thinking Process (With code interpreter's support):} \\
        \bottomrule
    \end{tabular}
\end{table*}

\noindent
\textbf{Prompt templates used in RL training.}
We provide the RL training prompt template in \textbf{\Cref{tab:rl_prompt}}. This template illustrates the input–output format and executable code constraints used during RL rollouts, offering additional transparency and reproducibility of our training setup.

\begin{figure*}[ht]
    \centering
    \includegraphics[width=1\linewidth]{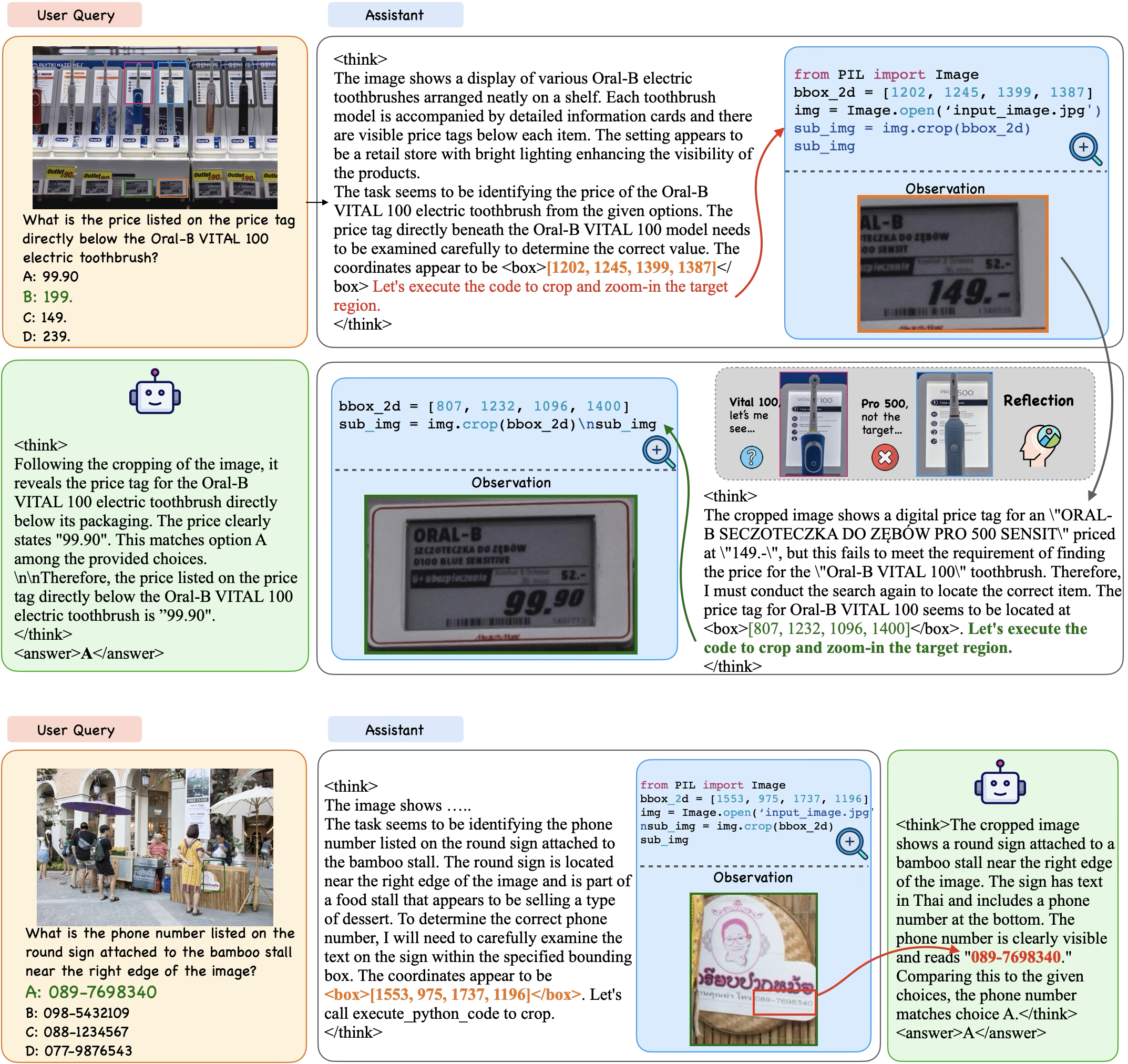}
    \caption{
    Example SFT training data of an atomic operation (\texttt{zoom}-\texttt{in} operation here). Top: two-turn, bottom: a single-turn trajectory.
    }
    \label{fig:sft-example}
\end{figure*}

\noindent
\textbf{Example SFT training data.}
To better illustrate the construction of SFT data, we provide representative examples of atomic operations. As shown in \Cref{fig:sft-example}, the top trajectory corresponds to a two-turn reasoning process, where the model iteratively performs cropping, observes intermediate results, and reflects on the correctness before locating the accurate price tag of a specific toothbrush. In contrast, the bottom trajectory demonstrates a single-turn process, in which the model directly identifies the phone number from a cropped sign. These cases exemplify how SFT data captures both multi-step and single-step reasoning, integrating tool invocation, visual observation, and answer generation.

\begin{table}[t]
    \centering
    \caption{Prompt template for Reinforcement Learning Rollout.}
    \label{tab:rl_prompt}
    \setlength{\tabcolsep}{6pt}
    \renewcommand{\arraystretch}{1.15}
    \begin{tabular}{p{0.95\linewidth}}
        \toprule
        \textbf{User.} \verb|<image>| Question: \verb|{question}| \\
        Think step-by-step within \verb|<think></think>|. You now have the ability to selectively write executable Python code to enhance your reasoning process. The Python code should be complete scripts, including necessary imports. \\
        Each code snippet is wrapped with \\
        \verb|<code>|\\
        \verb|```python|\\
        \verb|code snippet|\\
        \verb|```|\\
        \verb|</code>|\\[2pt]\\
        You must provide your final answer in \\\verb|<answer> </answer>|. \\
        \bottomrule
    \end{tabular}
\end{table}

\section{Additional empirical studies}

\subsection{Visualization of sequence/turn-level rewards}
\label{appendix:rbat_vis}

Figure~\ref{fig:sandbox_app} illustrates how the sequence-level reward $R_{\text{seq}}$ and the turn-level reward $R_{\text{turn}}$ defined in the main paper are applied along a multi-turn trajectory in the persistent execution environment. Each query is solved through a sequence of \emph{Reasoning–Code–Response} turns, where code is executed in a shared sandbox so that intermediate variables and visual artifacts can be reused across turns. After the whole trajectory $\tau$ is completed, the sequence-level reward $R_{\text{seq}}$ is computed from the final-answer correctness and the global tool-use statistics $(N_{\text{succ}}(\tau), N_{\text{total}}(\tau), \mu_{\text{acc}})$. The corresponding sequence-level advantage $A_{\text{seq}}$ is then broadcast to all tokens in the trajectory (green bars at the bottom of the figure), providing an outcome-level learning signal that adaptively encourages or discourages overall tool usage depending on task difficulty.

Inside the red dashed box, the turn-level component is computed from execution outcomes at each turn $m$. Failed executions receive an immediate negative penalty $r_{\text{turn},m}<0$, while successful or no-code turns receive $r_{\text{turn},m}=0$. These per-turn penalties are accumulated into discounted returns $G_{\text{turn}}^{m}$ and normalized into advantages $A_{\text{turn}}^{m}$, which are only assigned to the tokens of that specific turn (colored segments in the figure; hatched regions indicate zero turn-level signal). The final token-wise advantage is obtained by combining the global $A_{\text{seq}}$ with the local $A_{\text{turn}}^{m}$ of its turn.

\subsection{Hyperparameters in our method}
\label{appendix:hyperparams}
\subsubsection{Sensitivity of $R_{BAT}$ Hyperparameters}

\begin{figure}
    \centering
    \includegraphics[width=1\linewidth]{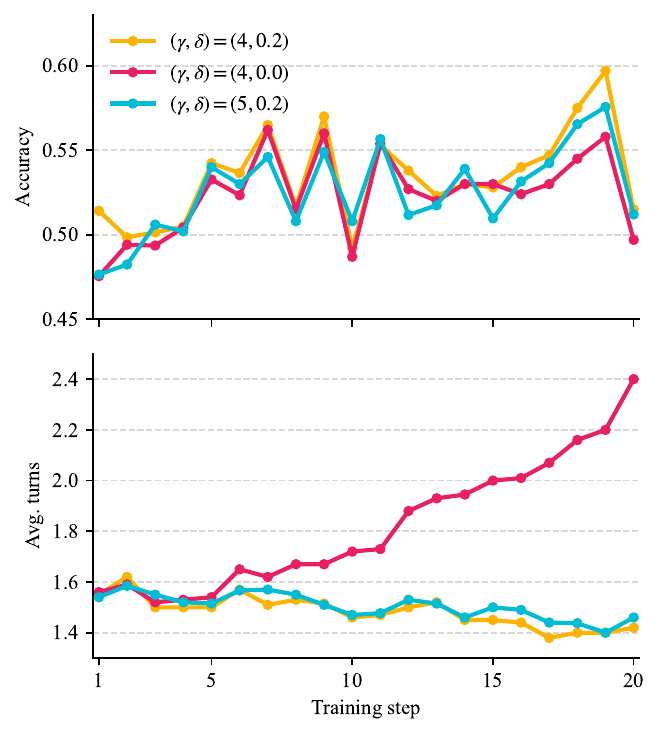}
    \caption{
Training dynamics of $R_{BAT}$ under three $(\gamma,\delta)$ configurations. 
We plot accuracy (top) and average reasoning turns (bottom) over training steps on the union of RL benchmarks. 
Accuracy is relatively stable across $\gamma\in\{4,5\}$ (i.e., $\gamma$ mainly smooths the transition in $d(\mu_{\text{acc}})$), whereas setting $\delta=0$ removes suppression for high-$\mu_{\text{acc}}$ groups and leads to steadily increasing tool usage with only marginal accuracy gains. 
Our default $(\gamma,\delta)=(4,0.2)$ maintains high accuracy while restraining unnecessary tool calls.
}
    \label{fig:rbat_gamma_delta_sweep}
\end{figure}

Recall that the sequence-level component of our reward is defined in Eq.~\eqref{eq:rseq} as
\begin{equation}
R_{\text{seq}}(\tau)
= \Bigl(0.5 + 0.5\cdot \mathbf{1}\{R_{\text{acc}}(\tau)>0\}\Bigr)
\, d(\mu_{\text{acc}})
\, \frac{N_{\text{succ}}(\tau)}{N_{\text{total}}(\tau)},
\label{eq:rseq}
\end{equation}
where the scaling term $d(\mu_{\text{acc}})$ is defined as $d(\mu_{\text{acc}}) = \sigma\!\bigl(\gamma(0.5-\mu_{\text{acc}})\bigr) - \delta$ , $\mu_{\text{acc}}\in[0,1]$ is the group-level accuracy, $\sigma(z)=1/(1+e^{-z})$ is the logistic function, and $(\gamma,\delta)$ control how strongly the group accuracy modulates the incentive on tool calls. Here, $\frac{N_{\text{succ}}(\tau)}{N_{\text{total}}(\tau)}$ measures the proportion of successful tool executions within the trajectory, and the leading term $\bigl(0.5 + 0.5\cdot \mathbf{1}\{R_{\text{acc}}(\tau)>0\}\bigr)$ switches the base weight according to final-answer correctness. Intuitively, $d(\mu_{\text{acc}})$ decreases as $\mu_{\text{acc}}$ increases: it amplifies successful tool calls when the group is struggling (low $\mu_{\text{acc}}$) and suppresses unnecessary calls when the group already performs well (high $\mu_{\text{acc}}$). The hyperparameters control this behavior: $\gamma$ adjusts how sharp the transition is, $\delta$ shifts the baseline and determines whether suppression occurs in high-accuracy regimes, and $\beta$ is the discount factor for per-turn penalties in $R_{\text{turn}}$. In the main experiments we set $\gamma=4$, $\delta=0.2$, and $\beta=0.2$. The sensitivity results are summarized in \Cref{fig:rbat_gamma_delta_sweep}.

\subsubsection{Effective range and bounds of $d(\mu_{\text{acc}})$}
As $\sigma(\cdot)\in(0,1)$ for all real inputs and $\mu_{\text{acc}}\in[0,1]$, the scaling term $d$ is deterministically bounded as
\[
-\delta < d(\mu_{\text{acc}}) < 1 - \delta.
\]
Importantly, negative values of $d$ are intentional rather than pathological: they occur only when the group-level accuracy $\mu_{\text{acc}}$ is sufficiently high and are used to discourage unnecessary tool calls on easy queries. The sign flip of $d$ happens at a unique threshold
\[
\mu_{\text{acc}}^\star = 0.5 - \frac{1}{\gamma}\,\sigma^{-1}(\delta),
\quad \text{where } \sigma^{-1}(u)=\log\frac{u}{1-u}.
\]
For $\gamma=4$, $\delta=0.2$, this gives a transition around
\[
\mu_{\text{acc}}^\star \approx 0.5 + \frac{1}{4}\log\frac{1-0.2}{0.2},
\]
i.e., only when most rollouts in the group are already correct does $d$ become slightly negative. In all other regimes (medium or low $\mu_{\text{acc}}$), $d>0$ and $R_{BAT}$ increases the relative value of successful tool calls. Together with the per-turn penalty $R_{\text{turn}}$ for failed execution, this design prevents degenerate tool-spamming or tool-off behaviors: on easy tasks, negative $d$ discourages extra calls; on hard tasks, positive $d$ amplifies the benefit of correctly executed tools.
\subsubsection{Hyperparameter over $(\gamma,\delta)$}
We run an ablation on $(\gamma,\delta)$ to probe $R_{BAT}$'s behavior in practice. Concretely, we train three settings, $(4,0.2)$, $(4,0.0)$ and $(5,0.2)$, and monitor training accuracy and average tool calls per example. This lets us (i) compare $\gamma\in{4,5}$ at fixed $\delta=0.2$, and (ii) isolate the effect of enabling vs.\ disabling suppression at high group-level accuracy by varying $\delta\in{0,0.2}$ at fixed $\gamma=4$.

Empirically, $\gamma$ has little effect on training accuracy, while setting $\delta=0$ removes suppression for high-$\mu_{\text{acc}}$ groups and causes substantially more tool calls with similar accuracy, i.e., tool overuse. Our choice $(4,0.2)$ keeps accuracy high while restraining unnecessary tool usage. The result is shown as \Cref{fig:rbat_gamma_delta_sweep}.

\subsection{Analysis of Reasoning Depth}
\label{appendix:reasoning_depth}

\begin{figure}[h]
    \centering
    \includegraphics[width=0.95\linewidth]{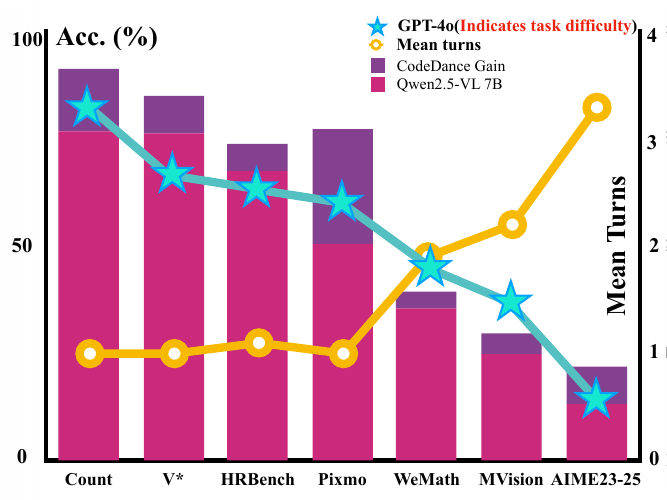}
    \caption{Reasoning depth vs. task difficulty.}
    \label{fig:reasoning_depth_p2}
\end{figure}

\textbf{Our policy aligns reasoning depth with task difficulty.} We plot \textit{task difficulty} (measured by GPT-4o ACC.) vs. \textit{number of turns} (see Fig.~\ref{fig:reasoning_depth_p2}) and validate that harder problems indeed induce longer reasoning turns, while easier ones terminate early.

\section{Additional qualitative results}
\label{appendix:qualitative}

\begin{figure*}[t]
        \centering
        \includegraphics[width=1\linewidth]{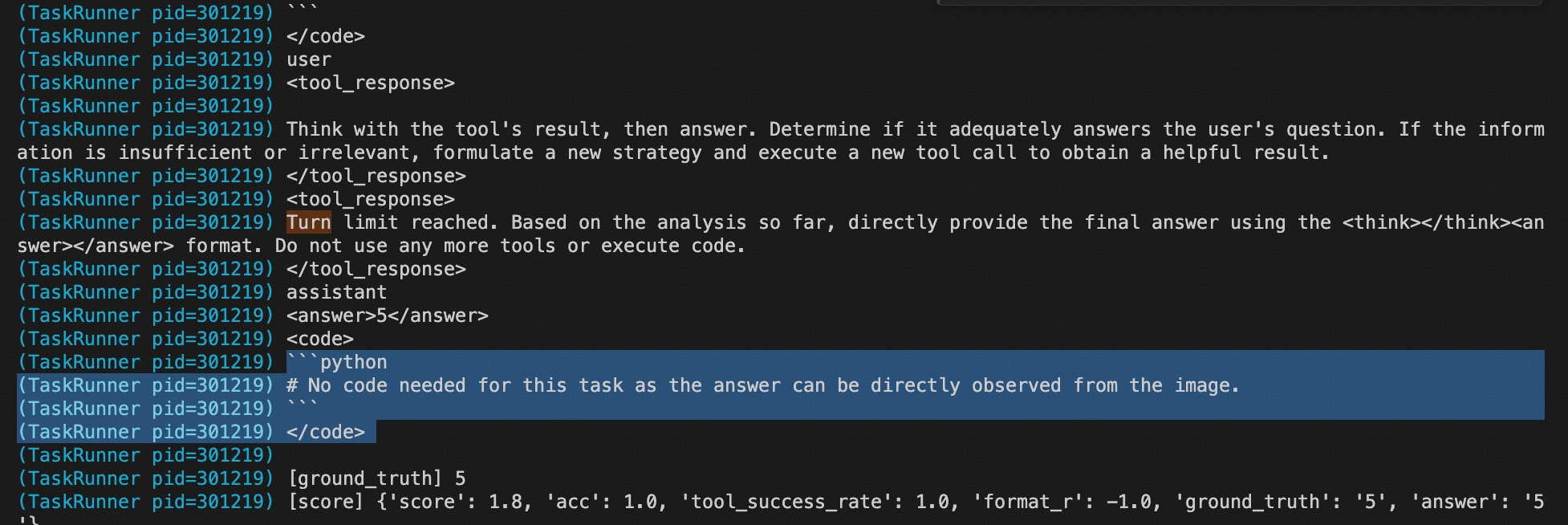}
        \caption{An example of a reasoning trajectory exhibiting reward hacking (using a DeepEyes-style tool reward for code generation). The MLLM exploits the reward by generating code with only commentary lines, which is not actually executed.}
        \label{fig:reward_hacking}
\end{figure*}

\textbf{Reward-hacking case when using naive tool call reward for code generation.}
A naive reward scheme that simply reinforces every successful tool call is prone to reward hacking, where the model exploits loopholes in the reward design rather than genuinely improving reasoning. For instance, we observe failure cases in Figure \ref{fig:reward_hacking} in which the model generates degenerate tool outputs (e.g., code consisting only of commentary lines without actual execution) that nevertheless satisfy superficial reward signals. Such behaviors artificially inflate tool success metrics while providing no real contribution to solving the task, thereby misleading training and undermining reasoning quality.

\subsection{Failure cases observed in our experiments} 
\begin{figure*}[h]
    \centering
    \includegraphics[width=1\linewidth]{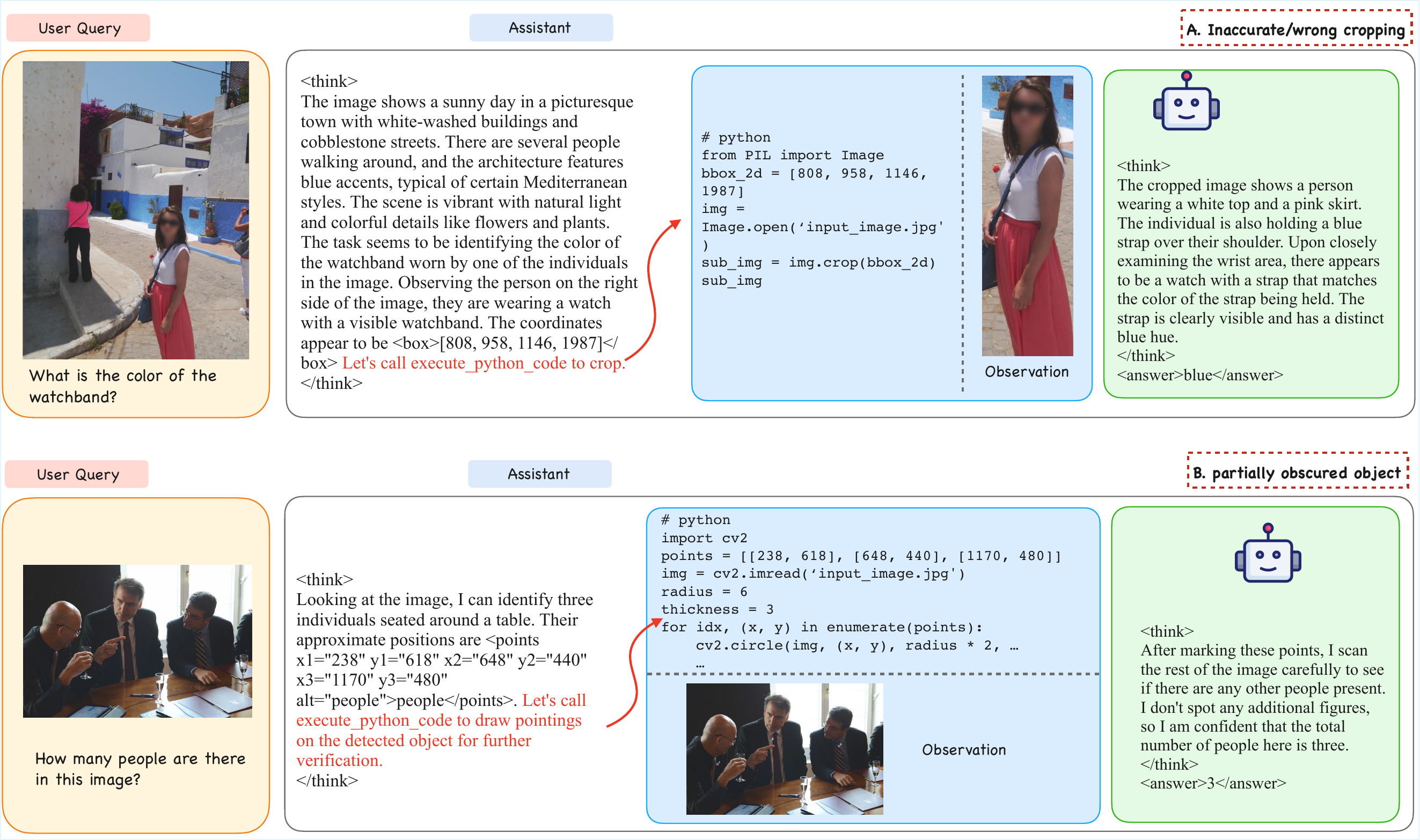}
    \caption{
    Failure cases found in our empirical studies. \textbf{A:} The model performs wrongly cropping. 
    \textbf{B:} The person on the right edge is partially obscured, thus hard to count.}
    \label{fig:failure_case}
\end{figure*}

Although our model performs competitively across benchmarks, several failure modes are still observed, and we summarize them in \Cref{fig:failure_case}.

First, inaccurate or suboptimal cropping may occur when the model misidentifies the target region or when visual cues are subtle. 
\textbf{As shown in Case A}, the model selects an overly large bounding box around the person, causing the cropped image to mix multiple objects; the model still answers correctly, but the tool invocation is clearly misaligned. Second, the model may miss partially obscured objects in complex scenes. 
\textbf{In Case B}, our model correctly identifies and marks visible individuals but fails to consider whether additional people might be hidden or partially occluded, leading to an incomplete final count. 

These cases highlight the remaining challenges in precise localization and robust multi-step verification under ambiguous or cluttered visual conditions. 


\subsection{Additional novel reasoning trajectories} 
\begin{figure*}[h]
    \centering
    \includegraphics[width=1\linewidth]{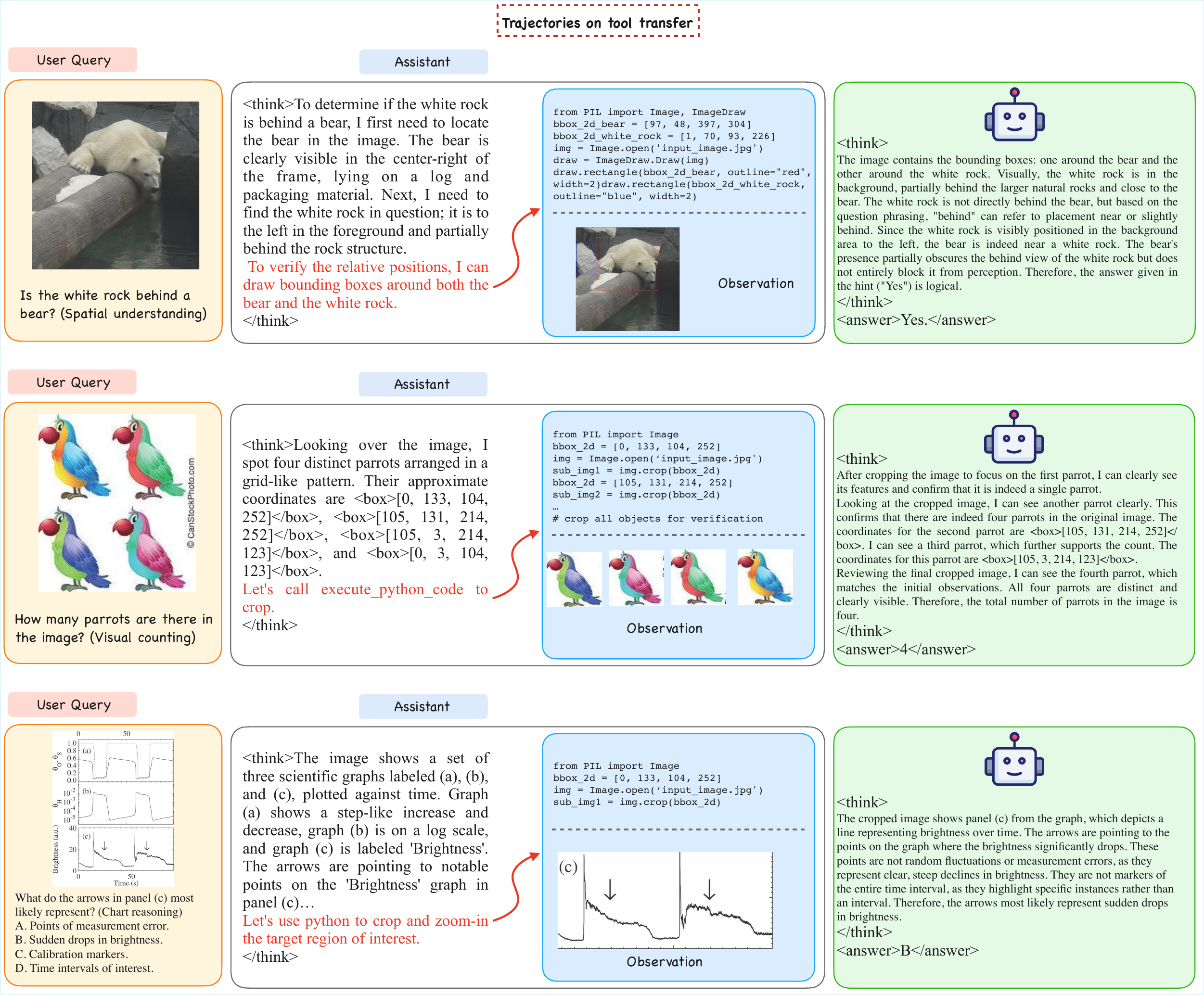}
    \caption{
    Novel reasoning trajectories on tool transfer to other tasks.
    }
    \label{fig:emergents_supp1}
\end{figure*}

In \Cref{fig:emergents_supp1}, we show step-by-step reasoning trajectories across three distinct vision tasks on tool transfer: 

\Cref{fig:emergents_supp1} (Top): The assistant tackles a spatial relational question by first localizing both the bear and the white rock using bounding boxes. It then uses PIL to draw red/blue rectangles around each object, visually verifying their relative positions. This demonstrates code-mediated spatial reasoning. Instead of relying on implicit attention maps or pretrained spatial priors, the model actively constructs visual evidence through code. The act of drawing bounding boxes serves as an internal ``visual scratchpad'', enabling explicit comparison of object positions, which is crucial for fine-grained spatial inference where ambiguity exists. 

\Cref{fig:emergents_supp1} (Mid): The assistant identifies four candidate regions containing parrots based on initial visual inspection. It then executes a loop of \texttt{img.crop(bounding box)} operations to isolate each region, visually confirming that each cropped area contains a unique, clearly distinguishable parrot. This iterative cropping and verification ensures no over- or under-counting. 
This exemplifies verification-driven counting. Rather than predicting a number directly (which risks hallucination or confusion with similar objects), the system uses tool-based segmentation to reduce the problem to a series of binary verifications (``Is this one a parrot?''). The modularity of PIL operations allows the model to treat counting as a compositional task — scaling naturally to more complex scenes.

\Cref{fig:emergents_supp1} (Bottom): Given a scientific graph with brightness vs. time, the assistant zooms into panel (c) using \texttt{img.crop()} to focus on the region with arrows. It observes sharp downward spikes in the curve at those points and infers they represent sudden drops in brightness, not measurement noise or calibration artifacts — based on the magnitude and shape of the dips.

\begin{figure*}[h]
    \centering
    \includegraphics[width=1\linewidth]{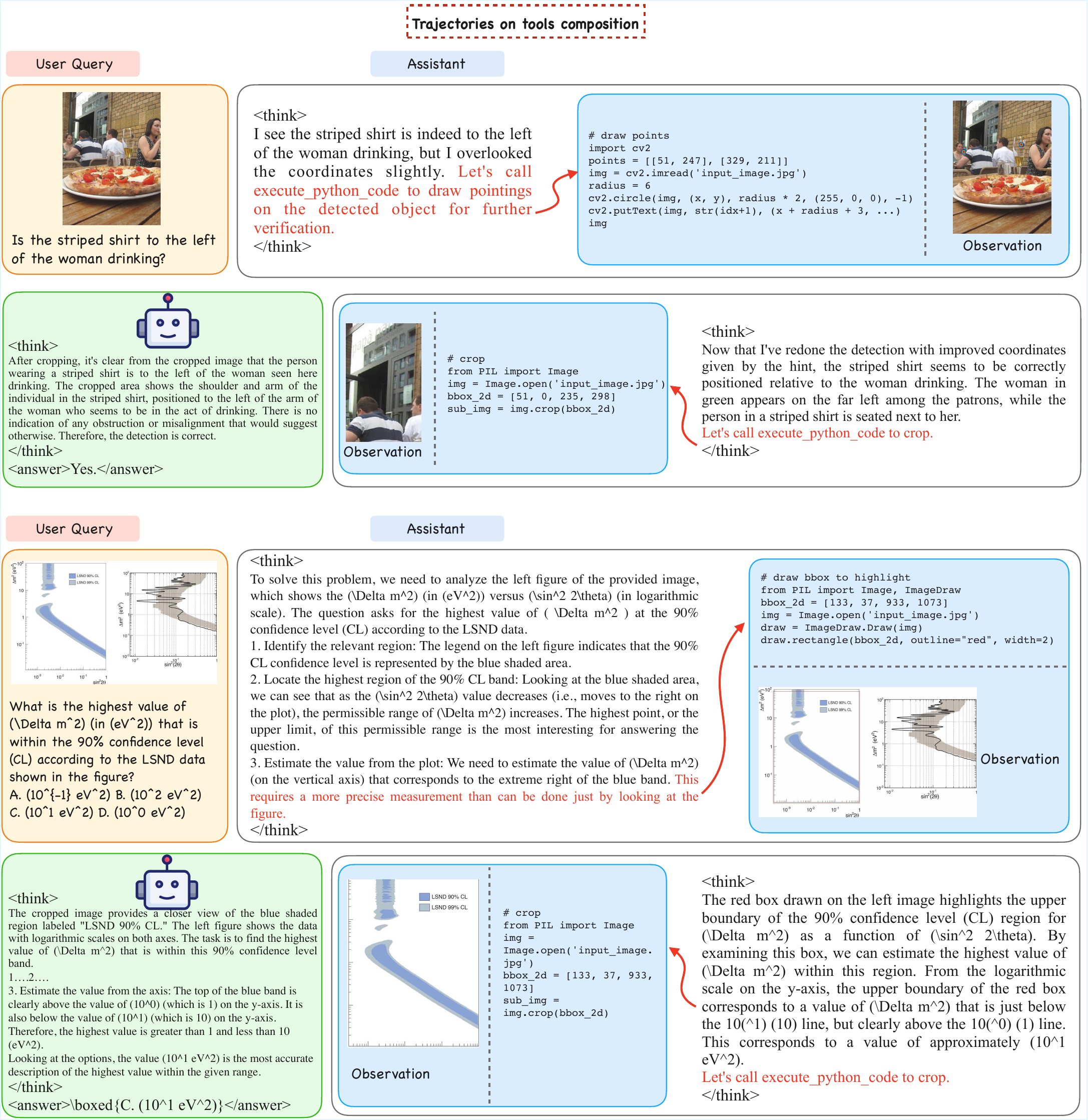}
    \caption{
    Reasoning trajectories on tools composition.
    }
    \label{fig:emergents_supp2}
\end{figure*}

Similarly, the trajectories in \Cref{fig:emergents_supp2}  reveal iterative, self-correcting reasoning enabled by dynamic tool composition, which is based on tool transfer ability since we only define a single set of tool/ability for each task during SFT. 

Top-row of \Cref{fig:emergents_supp2}: The assistant first attempts to locate the person in the striped shirt relative to the woman drinking. It initially misidentifies coordinates, so it composes two tools:
First, it uses \texttt{cv2.circle()}to draw red points at hypothesized locations — visually flagging potential errors.
Then, it corrects the coordinates and uses \texttt{PIL.Image.crop()} to zoom into the region for closer inspection.
Finally, it confirms the spatial relationship: the striped-shirt person is indeed to the left, seated next to the drinking woman — no occlusion or misleading posture.

Bottom-row of \Cref{fig:emergents_supp2}: The assistant must extract a precise numerical value from a scientific plot showing $(\Delta m^2)$ vs. $sin^2(2\theta)$. It follows a multi-step strategy:
Identify region: Uses \texttt{ImageDraw.rectangle()} to highlight the blue shaded 90\% confidence level (CL) band.
Zoom in: Crops the upper boundary of this region using \texttt{PIL.Image.crop()} to isolate the extreme right edge — where $(\Delta m^2)$ reaches its maximum within the CL.
Finally interpret scale and answer.

While these reasoning trajectories during RL exploration are not without flaws, e.g. occasionally exhibiting imprecise coordinate estimation or redundant tool calls, they collectively demonstrate the potential of tool-augmented multimodal reasoning.

\section{Discussion}
\noindent
\textbf{Why code-based tool use, rather than API-style calls?}
We adopt Python code as the medium for tool use because it provides a general and compositional interface. Unlike fixed API schemas, code naturally supports both tool invocation and program logic (e.g., sequencing, conditionals, loops, numerical computation). This richer interface allows models to flexibly define and combine operations, and it produces transparent and verifiable execution traces that can be systematically inspected. In practice, code also makes extension straightforward: adding a new tool only requires exposing its API, without redesigning templates, retraining connectors, or engineering complex prompts.

\noindent
\textbf{Why a single dense model, rather than an agent pipeline?}
A unified dense model offers several practical advantages over modular agent workflows:
(1) it avoids error propagation across multiple components by learning an end-to-end interface;
(2) it achieves lower latency and compute cost, since reasoning and tool orchestration are handled in a single forward pass;
(3) it is more robust, as performance does not hinge on the reliability of each sub-module; and
(4) it benefits from a unified optimization target, whereas agent systems often require additional policies or connectors to be separately tuned.

In addition, given realistic compute constraints, most of our experiments in this work are conducted with 7B-scale models (e.g., Qwen-2.5-VL-7B), where we already observe promising effects: consistent gains across general understanding and complex reasoning benchmarks, and the emergence of new behaviors (e.g., novel tool use and tool compositions of atomic skills to new tasks). 
These empirical observations \textit{are easier to scale} within a single dense model, while agent pipelines introduce many interacting modules that complicate both training and deployment.
Overall, our design favors simplicity, efficiency, and scalability, making it a more practical foundation for future progress.

\noindent

\section{Broader impact}
This work contributes toward building more transparent and verifiable multimodal reasoning systems by adopting executable code as the unified medium for tool use. 
The ability to generate interpretable traces and intermediate artifacts can benefit applications where accountability and auditability are essential, such as scientific analysis and education. 
At the same time, code-generating models pose risks: malicious users could potentially exploit them for unsafe automation, and generated visual artifacts might be misused to mislead or manipulate. 
To mitigate these concerns, we recommend pairing such systems with appropriate safeguards, including safety filters, usage constraints, and responsible deployment practices. By doing so, the benefits of executable visual reasoning can be realized while minimizing the potential for misuse.

\section{Limitations and future work}

\noindent
\textbf{Limitations.}
While our method demonstrates promising emergent behaviors and strong performance across diverse visual reasoning tasks, several limitations remain. First, the reliance on high-quality synthetic trajectories implies that certain real-world reasoning patterns may be underrepresented, potentially limiting robustness in open-domain scenarios. Second, although code provides a general interface, extending to richer modalities (e.g., audio) or domain-specific tools (e.g., medical applications) will require additional engineering. Finally, due to compute constraints, our evaluations are primarily conducted on 7B-scale models; the scalability of emergent behaviors at larger scales remains to be systematically examined. 
Nevertheless, our preliminary experiments suggest a promising trend when scaling up model capacity and compute resources.

\clearpage
\noindent
\textbf{Future Work.}
Our framework demonstrates the potential of multimodal reasoning models to support natural conversations with seamless and proactive tool use through executable code, thereby enabling more advanced problem-solving capabilities. 
Looking ahead, we envision that the ability to ``think with images'' will evolve beyond the vision modality and fixed schemas, fostering novel tool discovery and the spontaneous composition of tools in a more generalized and efficient manner. 
Such directions may ultimately pave the way toward multimodal agents that are both versatile and adaptive across diverse domains.




